\documentclass[10pt,journal,compsoc]{IEEEtran}
\ifCLASSOPTIONcompsoc
  \usepackage[nocompress]{cite}
\else
  \usepackage{cite}
\fi
\usepackage{amssymb}
\usepackage{url}
\usepackage{multirow}
 \usepackage{amsmath}
\usepackage{array}
\usepackage{bm}
\usepackage{subcaption}
\usepackage{graphicx}
\usepackage{color, colortbl}
\definecolor{LightCyan}{rgb}{0.88,1,1}
\definecolor{LightPurple}{rgb}{0.79607843137,0.80784313725,0.98431372549}
\ifCLASSINFOpdf
\else
\fi
\begin{document}
%
\title{OpenVeinNet: Robust Open-Set Finger Vein Verification with Dynamic Snake Convolution and Graph Learning}
%
%
%
%

\author{Sushrut Patwardhan and
        Raghavendra Ramachandra
\thanks{Both authors have contributed equally. \\
SAFE Centre, Department of Information Security and Communication Technology (IIK), Norwegian University of Science and Technology (NTNU), Gjøvik, Norway.  \\
            Corresponding Author: Raghavendra Ramachandra (E-Mail: raghavendra.ramachandra@ntnu.no)}}

\IEEEtitleabstractindextext{%
\begin{abstract}
Finger vein verification is a promising biometric modality for secure
authentication because vascular patterns are internal, difficult to observe
externally, and relatively resistant to presentation attacks. However, reliable
verification remains challenging in open-set settings, where test identities are
unseen during training and non-enrolled probes must be rejected at inference.
This paper presents OpenVeinNet, a finger vein verification framework designed
for cross-dataset and open-set evaluation. The proposed model combines Dynamic
Snake Convolution with graph-based feature modelling. Dynamic Snake
Convolution extracts local curvilinear and tubular vein structures using
adaptive sampling, while the graph convolutional backbone models long-range
topological relationships between vein regions. To improve the discriminative
quality of the embedding space, we introduce a Centroid Angular Hybrid Loss,
which jointly encourages intra-class compactness and inter-class angular
separation for cosine-similarity-based verification. Experiments are conducted
on five public finger vein datasets: FV-300, MMCBNU, FV-USM, PolyU, and VERA.
The method is evaluated using leave-one-dataset-out training under both
enrolment-based unknown-rejection and full-subject verification protocols, and
is compared with handcrafted and recent deep learning-based baselines. The
results show that OpenVeinNet achieves strong cross-dataset generalisation,
consistently low equal error rates, and competitive true accept rates at fixed
false accept rate operating points. Ablation studies further confirm the
individual and combined contributions of adaptive tubular feature extraction,
graph-based relational modelling, and the proposed loss function. These findings
indicate that explicitly modelling local vein geometry, global vascular
relationships, and angularly compact embeddings is effective for open-set finger
vein verification.
\end{abstract}

}

\maketitle

\IEEEdisplaynontitleabstractindextext
\IEEEpeerreviewmaketitle

\section{Introduction}

Reliable user verification is essential for secure access control systems across a wide range of applications. Biometric authentication has become a widely adopted solution because it offers convenient and identity-linked access control. Conventional biometric traits such as fingerprints, facial features, irises, palmprints, and ears are commonly used; however, they may be vulnerable to spoofing, particularly when acquired in a non-intrusive manner. In contrast, finger vein patterns provide an attractive alternative because the vascular structures are located beneath the skin and are therefore difficult to observe externally or replicate. This intrinsic resistance to presentation attacks, together with high verification accuracy and relatively low deployment cost, has increased the use of finger vein recognition, particularly in banking and financial services.

Finger vein patterns are typically acquired using cost-effective near-infrared (NIR) imaging systems, where the light source and camera are positioned above, alongside, or around the finger. When NIR light penetrates the finger, it is absorbed by haemoglobin in the veins, causing the vascular structures to appear darker than the surrounding tissue. Finger vein biometrics are also characterised by temporal stability, relatively low sensitivity to external environmental variations, and strong uniqueness across individuals. Furthermore, capturing images from both the dorsal and palmar sides of the finger can improve verification performance.

\begin{figure}[ht!]
\centering
\includegraphics[width=\linewidth]{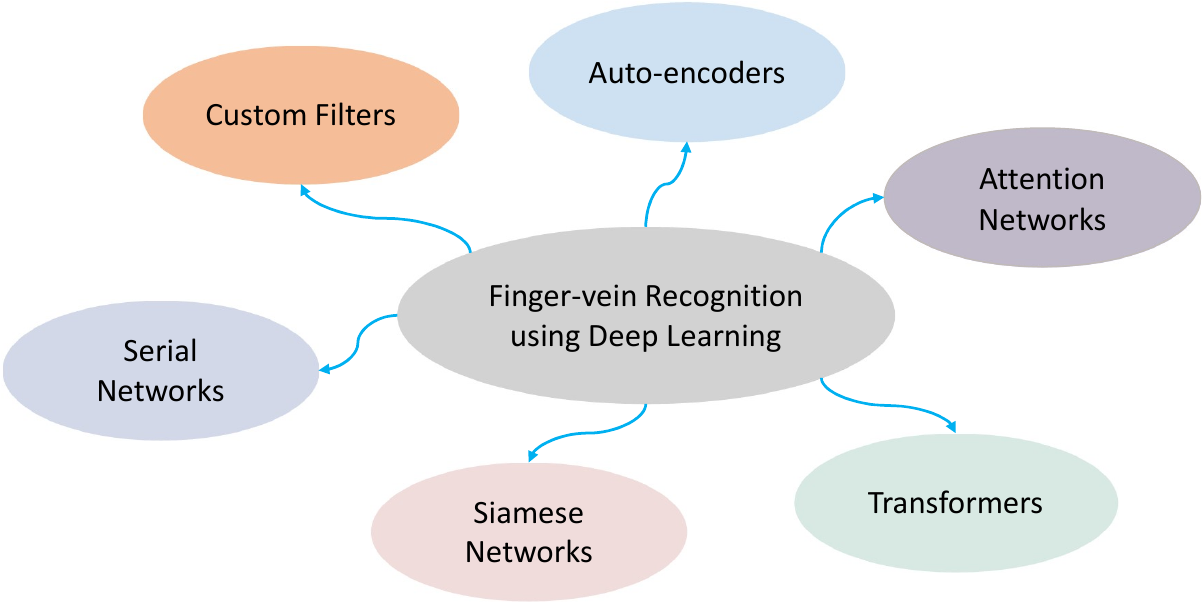}
\caption{A brief taxonomy of finger vein verification techniques based on deep learning approaches.}
\label{fig:techniques}
\end{figure}

\begin{table*}[hbt!]
    \centering
    \resizebox{\textwidth}{!}{
        \begin{tabular}{|l|p{4in}|}
            \hline
             \rowcolor{LightPurple} Authors (Year) & Technique Description   \\
              \hline
                Huafeng Qin et al., \cite{2015_1} (2015) & 3 Sequential convolution layers and 2 dense layers \\ \hline
                Itqan et al., \cite{2016_1} (2016) & 3 Sequential convolution layers and 1 dense layers \\ \hline
                Syafeeza Radzi et al., \cite{2016_2} (2016) & 2 Sequential convolution layers\\ \hline
                Huafeng Qin et al., \cite{2017_1} (2017) & 4 Sequential convolution layers with path-based training.\\ \hline
                Hyung Gil Hong et al., \cite{2017_2} (2017) & 12 Sequential convolution layers and 3 dense layers\\ \hline
                 Rig Das et al., \cite{2018_1} (2018) & 5 Sequential convolution layers\\ \hline
                Cihui Xie et al., \cite{2019_1} (2019) & Siamese network with 5 conventional layers trained on triplet loss function.\\ \hline
                Jong Min Song et al., \cite{2019_2} (2019) & 8 Sequential convolution layers trained using composite finger vein image generated by converting the 1-channel input image to a 3-channel input image.\\ \hline
                Su Tang et al., \cite{2019_3} (2019) & Siamese network with residual CNN architecture.\\ \hline
                Borui Hou et al., \cite{2019_4} (2019) & Autoencoder architecture.\\ \hline
                Wenming Yang et al.,\cite{2019_5} (2019) & Using GAN for generating enhanced images and performing verification using encoder features of GAN. \\ \hline
                Junying Zeng et al., \cite{2020_1} (2020) & Deformable convolution in U-NET type architecture.\\ \hline
                Ridvan Salih Kuzu et al., \cite{2020_2} (2020) & 6 Sequential convolution layers and 2 dense layers and LSTM for classification.\\ \hline
                Zhiang Hao et al., \cite{2020_3} (2020) & Multitask Learning based on Resnet-18 like model with SoftL1Loss and ArcFace loss\\
                \hline
                Hengyi Ren et al., \cite{2021_1} (2021) &  Squeeze and Excitation blocks embedded Resnet trained on encrypted finger vein images.\\ \hline
                Ridvan Salih Kuzu et al., \cite{2021_2} (2021) & Custom DenseNet 161 trained on additive angular penalty and large margin cosine penalty loss function. \\ \hline
                Borui Hou et. al., \cite{arcvein} (2021) & Resnet-like architecture with ECA-Resnet blocks trained on arc-cosine loss and softmax loss together.\\ \hline
                Zhenxiang Chen et. al., \cite{chenresnetarcloss} (2021) & Resnet-18 architecture trained on arc-loss with image enhancement, and luminance-inversion data augmentation\\ \hline
                Weili Yang et al., \cite{2022_1} (2022) & Multi-view finger vein with individual CNNs and view pooling. \\ \hline
                Huafeng Qin et al., \cite{2022_2} (2022) & Based on Local Attention Transformer\\ \hline
                Tingting Chai et al., \cite{2022_3} (2022) &  5 Sequential convolution layers and 1 dense layers \\ \hline
                Ismail et al., \cite{ism} (2022) &  Deeply-fused 3 Sequential convolution layers and 2 dense layers. \\ \hline
                Weiye Liu et al., \cite{2023_1} (2023) & Inception architecture with residual attention block.\\ \hline
                Zhongxia Zhang et al., \cite{2023_2} (2023) & CNN with spatial and channel attention module.\\ \hline
                Chunxin Fang et al., \cite{2023_3} (2023) & Siamese network with attention module.\\ \hline
                Chih-Hsien Hsia et al. \cite{2023_4} (2023) & Lightweight Convolutional Neural Network with attention and Margin based loss function.\\ \hline 
                Yiwei Huang et al. \cite{2023_5} (2023) & Axially enhanced local attention network (criss-cross kernel) \\ \hline 
                Huang, Junduan et al. \cite{2023_6} (2023) & Frequency spatial coupling network with resnet like architecture \\ \hline
                Bin Wa et al., \cite{binma} (2023) & 3 Sequential convolution layers with bilinear pooling and multiple attention modules.\\ \hline
                Xiaoye Li et al., \cite{vit} (2023) & Vision Transformer with rigorous regularisation added in mlp.\\ \hline
                Raghavendra Ramachandra et al., \cite{veinatnnet} (2024) & 3 Sequential convolution layers followed by multi-head attention module connected in parallel with normal and enhanced finger vein images.\\  \hline 
                Huafeng Qin et al., \cite{2024_2} (2024) & Generates optimal network using Neural Architecture Search using Gated Recurrent Units with self-attention. \\ \hline
                Pengyang Zhao et. al. \cite{2024_1} (2024) & Transformer-based view finger vein recognition model which uses enhanced images as attention mask\\ \hline
                Enyan Li et. al. \cite{lgfin} (2025) & Transformer and convolution-based model processing local and global features.\\ \hline
                \hline
                \textbf{Proposed Method} (\textbf{2026}) & \textbf{Dynamic Snake Convolution with Graph Convolution layers for feature extraction and novel Centroid-Angular Hybrid Loss.} \\
              \hline          
        \end{tabular}
    }
    \medskip
    \caption{Summary of existing deep learning-based finger vein verification techniques.}
    \label{tab:prevwork}
\end{table*}

\begin{figure*}[htp]
\centering
\includegraphics[width=1\textwidth]{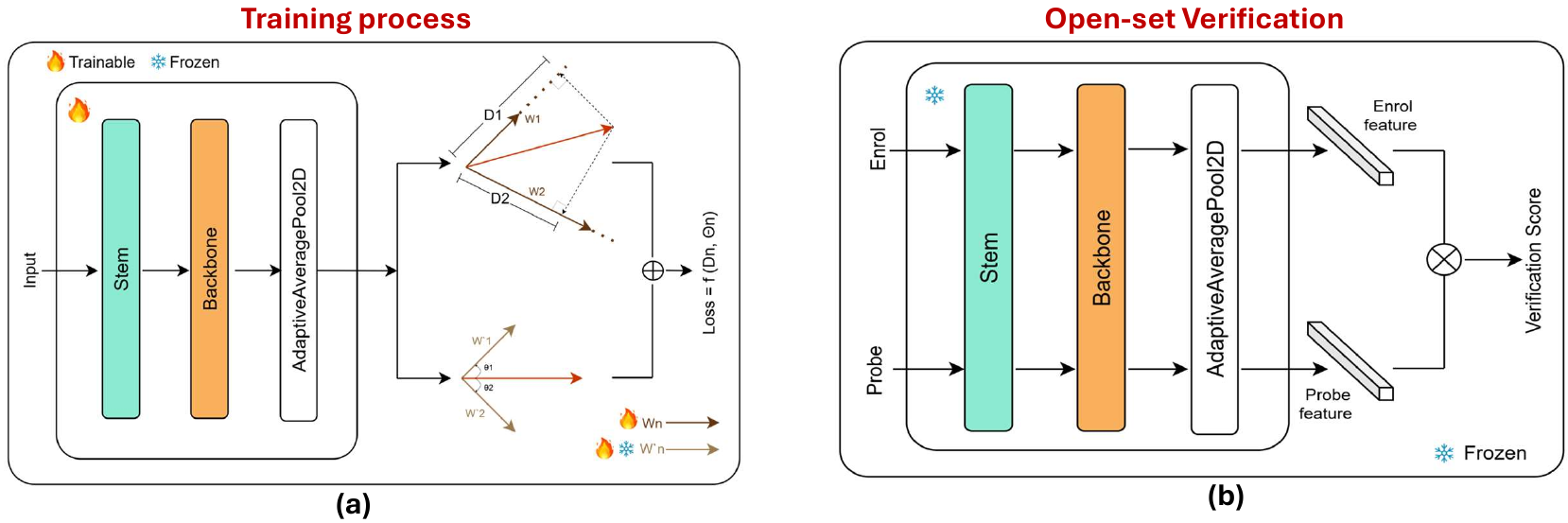}
\caption{Block diagram of the proposed OpenVeinNet framework for open-set finger vein verification. (a) The training phase incorporates a stem module with Dynamic Snake Convolution, followed by a multi-layer Graph Convolutional Network (GCN) and the proposed Centroid Angular Hybrid Loss for robust representation learning. (b) During inference, the system performs open-set verification by comparing probe samples against enrolled templates using the learned graph-based embeddings.}
\label{fig:model}
\end{figure*}

Finger vein verification has been extensively investigated in the literature, resulting in a wide range of handcrafted and deep learning-based methods. Earlier studies primarily relied on image processing techniques to extract vein patterns for template-based matching. Representative handcrafted methods include Maximum Curvature Points (MCP)~\cite{mcp}, Repeated Line Tracking (RLT)~\cite{rlt}, Wide Line Detectors (WLD)~\cite{wld}, and Mean Curvature (MC), which enhance vascular structures and can achieve reliable verification under controlled acquisition conditions. Texture descriptors such as Local Binary Pattern (LBP)~\cite{FV_LBPV}, Local Line Binary Pattern (LLBP)~\cite{rosdi2011finger}, Histogram of Oriented Gradients (HoG)~\cite{FV_HoG}, and Local Directional Pattern (LDP) have also been explored. In addition, dimensionality reduction methods, including Principal Component Analysis (PCA) and Linear Discriminant Analysis (LDA), have been used to improve computational efficiency and reduce feature redundancy. Although these handcrafted approaches remain effective in clean and well-controlled scenarios, they often struggle to generalise across changes in image quality, capture devices, and acquisition environments~\cite{surveyonmanualfeatures}.

In recent years, deep learning has been extensively explored for finger vein verification, yielding strong performance gains over traditional methods. A taxonomy of deep learning algorithms for finger vein verification is illustrated in Figure~\ref{fig:techniques}, and a summary of representative techniques is provided in Table~\ref{tab:prevwork}. Early deep learning-based approaches typically employed serial convolutional neural network (CNN) architectures, consisting of convolutional layers followed by one or two fully connected layers. Such models, as used in~\cite{2015_1, 2016_1, 2016_2, 2017_1,2017_2, 2018_1, 2019_2, 2020_2, 2020_3, 2021_1, 2021_2,chenresnetarcloss,  2022_1, 2022_3, ism}, are generally lightweight and computationally efficient, but their generalisation can be limited when evaluated across unseen acquisition conditions or new subject populations.

Autoencoder-based networks have been used to learn high-level feature representations for finger vein verification~\cite{2019_4, 2019_5}. Deformable filters incorporated into Fully Convolutional Networks (FCNs) have also been proposed to better model spatial variations~\cite{2020_1}. While these approaches improve local adaptability, they may require longer training and can still be sensitive to sensor-specific characteristics. Siamese networks have also been investigated for finger vein verification~\cite{2019_1, 2019_3, 2023_3}. Their pairwise training formulation can help with class imbalance, but such models are computationally demanding to train and may become difficult to scale as the number of enrolled identities increases.

Attention-based architectures have attracted growing interest because they can model both local and global discriminative cues within a unified framework~\cite{arcvein, binma, 2022_2, 2023_1, 2023_2, 2023_3, 2023_4, 2023_5, vit, veinatnnet, 2024_1, 2024_2}. However, their performance can degrade under cross-dataset evaluation when the training and testing data differ in sensor characteristics, image quality, and acquisition protocols. Vision Transformer (ViT)-based methods have also been explored~\cite{2022_2, vit}, but their effectiveness is often constrained by the limited scale of publicly available finger vein datasets. More recently, LGFIN~\cite{lgfin} combines multi-scale local and global contextual features using a dynamic feature interaction network, showing the importance of jointly modelling complementary vein information.

Despite these advances, two challenges remain central for practical finger vein verification. First, many methods show reduced robustness under cross-dataset evaluation because the learned representations are affected by sensor and acquisition shifts. Second, open-set deployment requires not only matching enrolled identities but also rejecting non-enrolled probes at fixed operating thresholds. This distinction is important because a system that performs well when all test identities are enrolled may still fail when unknown identities are presented during inference \cite{chenresnetarcloss}.

To address these limitations, we propose \emph{OpenVeinNet}, a finger vein verification framework designed for robust cross-dataset and open-set evaluation. The proposed model uses Dynamic Snake Convolution (DSConv) layers to capture the curvilinear and tubular structures of finger vein patterns. These features are then processed by Graph Convolutional Networks (GCNs), which model long-range topological relationships between vein regions. For verification, the extracted embeddings are compared using cosine similarity between enrolment and probe samples. To further improve open-set discrimination, we introduce a Centroid Angular Hybrid Loss that encourages intra-class compactness and inter-class angular separation in the learned embedding space. 

We evaluate the proposed approach on five publicly available finger vein datasets, namely FV-300, FV-USM, MMCBNU, PolyU, and VERA, and compare it against handcrafted and deep learning-based state-of-the-art methods. The evaluation includes both enrollment-based unknown-rejection and full-subject verification protocols, together with threshold-based operating points such as TAR@FAR. The results show that OpenVeinNet achieves  overall generalisation and favourable verification performance across challenging cross-dataset settings. The main contributions of this work are as follows:

\begin{itemize}
    \item We introduce \emph{OpenVeinNet}, a finger vein verification framework that integrates DSConv-based tubular feature extraction with graph-based relational modelling to capture both local vein geometry and long-range vascular connectivity.
    \item We propose a Centroid Angular Hybrid Loss that improves discriminative representation learning by promoting intra-class compactness and inter-class angular separation, which is particularly beneficial for open-set verification.
    \item We conduct extensive experiments on five public datasets, including FV-300, FV-USM, MMCBNU, PolyU, and VERA, and benchmark the proposed method against handcrafted and deep learning-based state-of-the-art methods.
    \item We evaluate the method under both enrollment-based unknown-rejection and full-subject verification protocols, reporting AUC, EER, and threshold-based TAR@FAR operating points to reflect practical biometric deployment conditions.
    \item We provide ablation studies, computational-cost analysis, and Grad-CAM++-based interpretability analysis to justify the architectural design, loss formulation, and decision behaviour of the proposed model.
    \item The complete source code, trained checkpoints, and scripts required to reproduce the reported results are available at:
    \url{https://github.com/Blazkowiz47/deep-vein-gcn}
\end{itemize}

\noindent Beyond architectural design, the experiments yield three 
analytical observations relevant to finger vein verification. 
The component ablation shows that graph-based relational modelling 
is only effective when local node features are geometrically aligned 
with vein structures, revealing a prerequisite specific to tubular 
biometric patterns. The loss ablation shows that explicitly penalising intra-class scatter through centroid alignment provides a practical advantage over margin-only losses in cosine-similarity-based open-set 
verification.

The remainder of this paper is organised as follows. Section~\ref{sec:proposed} presents the proposed method, including its architectural components and loss function. Section~\ref{sec:experiments} reports the experimental protocols and quantitative comparison with state-of-the-art methods. Section~\ref{sec:ablation} presents the ablation studies used to validate the main design choices. Section \ref{sec:efficiency} discuss the computation cost and latency, Section~\ref{sec:interpretation} provides the interpretability analysis. Finally, Section~\ref{sec:conclusion} concludes the paper and discusses future research directions.
\section{Proposed Method: OpenVeinNet for Open-Set Finger Vein Verification}
\label{sec:proposed}

Finger vein recognition relies on the distinctive curvilinear patterns of veins located beneath the skin for biometric verification. Although these patterns are unique to each finger and individual, reliable verification remains challenging due to variations in vein visibility, image quality, finger pose, rotation, and acquisition conditions. These challenges become more pronounced in open-set verification, where evaluation identities are unseen during training and non-enrolled probe samples must be rejected at inference time. To address these issues, we propose \textit{OpenVeinNet}, a finger vein verification framework that combines a Dynamic Snake Convolution Graph Neural Network (DscGrapher) with a Centroid Angular Hybrid Loss. The DscGrapher is designed to model the tubular and branching structure of finger veins using Graph Convolutional Networks (GCNs)~\cite{deepgcns}. It first extracts vein-aware local features using Dynamic Snake Convolution (DSConv) \cite{dsc}, and then aggregates neighbouring regions into graph-based embeddings. The resulting compact feature representation is compared using cosine similarity for verification.

Figure~\ref{fig:model} shows the block diagram of the proposed \textit{OpenVeinNet}, and Table~\ref{tab:dims} provides the architectural details. Figure~\ref{fig:model}(a) illustrates the training phase, while Figure~\ref{fig:model}(b) shows the open-set verification phase. During inference, finger vein images are passed through \textit{OpenVeinNet} to extract discriminative embeddings. These embeddings are then compared using cosine similarity between the probe and enrolled reference samples.

As shown in Figure~\ref{fig:model}(a), \textit{OpenVeinNet} consists of two main components. The first is a stem module that extracts local tubular vein features using DSConv blocks. The second is a graph-based backbone composed of Grapher Blocks, which captures local and long-range relationships between vein regions. In addition to the architectural design, we introduce the Centroid Angular Hybrid Loss to encourage compact intra-class embeddings and improved inter-class angular separation. The stem module also includes a learnable positional embedding to preserve spatial structure before graph-based feature aggregation. The final global representation is obtained using adaptive average pooling and is used for verification.

The following subsections describe each component of the proposed \textit{OpenVeinNet} framework.

\subsection{Stem: Dynamic Snake Convolution (DSConv) Blocks}

Finger vein patterns form complex, non-repeating, branching structures. Standard convolutional layers operate on fixed and regular grids, which limits their ability to follow curved and irregular vascular patterns. To address this limitation, we employ Dynamic Snake Convolution (DSConv), which adjusts its sampling locations to better align with local vein contours. This adaptive sampling behaviour allows the network to extract features from faint, curved, or locally deformed vein structures more effectively than standard convolutions.
Unlike general deformable and orientation-adaptive convolutions, DSConv progressively constrains neighbouring sampling locations along a continuous curvilinear trajectory, providing a morphology-aligned inductive bias for following thin and tortuous finger-vein structures.

\begin{figure}[!ht]
\centering
\includegraphics[width=0.425\textwidth]{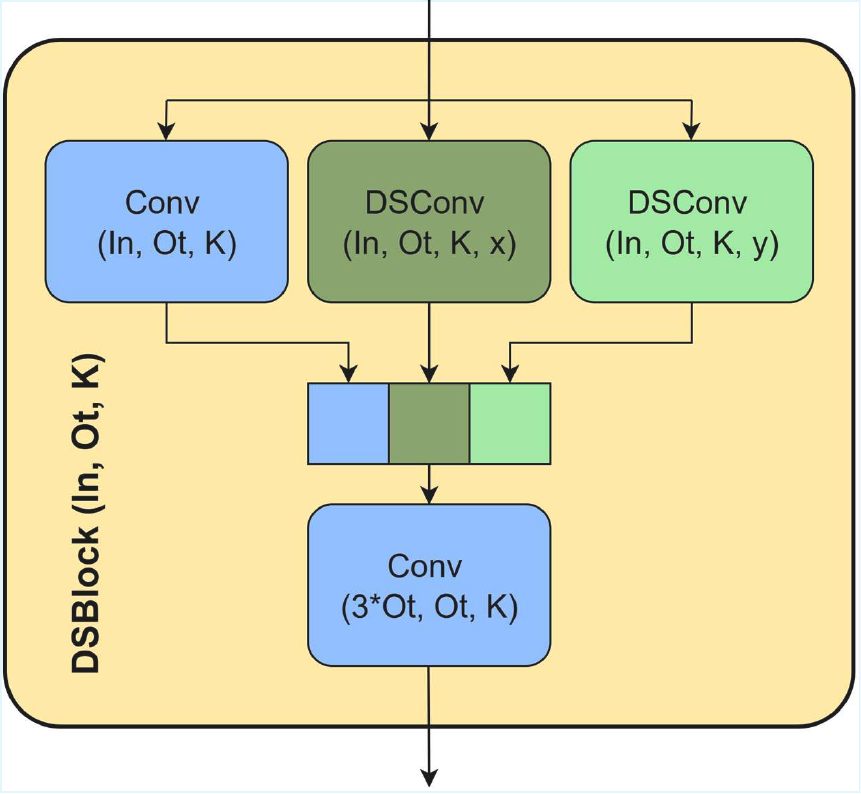}
\caption{DSBlock (In: input channels, Out: output channels, K: kernel size, x and y are the directions in which DSConv operates)}
\label{fig:dscblock}
\end{figure}

\begin{table}[hbt!]
    \centering
    \begin{tabular}{| c | c | c |}
         \hline
          Component & Layer & Output Dimensions \\
         \hline
         \hline
         Input & - & 224,224,3 \\
         \hline
                & DSC module 1 (S=2, F=3) & (112,112,16) \\
        Stem    & DSC module 2 (S=2, F=3) & (56,56,32) \\
                & DSC module 3 (S=2, F=3) & (28,28,64) \\
         \hline
                    & Stage I (K=18, S=2, F=3) & (14,14,128) \\
        Backbone    & Stage II (K=9, S=2, F=3) & (7,7,256) \\
                    & avg\_pool & (1,1,256) \\
         \hline
    \end{tabular}

    \medskip
    \caption{OpenVeinNet architecture details indicating output dimension. $K$ indicates the number of nearest neighbours, $S$ indicates stride, and $F$ indicates filter size.}
    \label{tab:dims}
\end{table}

Figure~\ref{fig:dscblock} illustrates the proposed DSConv Block. The block contains one standard convolution branch and two DSConv branches operating along the horizontal and vertical directions. The standard convolution preserves conventional local image evidence, while the DSConv branches adapt to direction-specific vein structures. The outputs of these branches are concatenated and passed through a final convolution to produce the block output. This design allows the stem to capture both regular local texture and deformable tubular features.

\begin{figure}[!ht]
\centering
\includegraphics[width=0.475\textwidth]{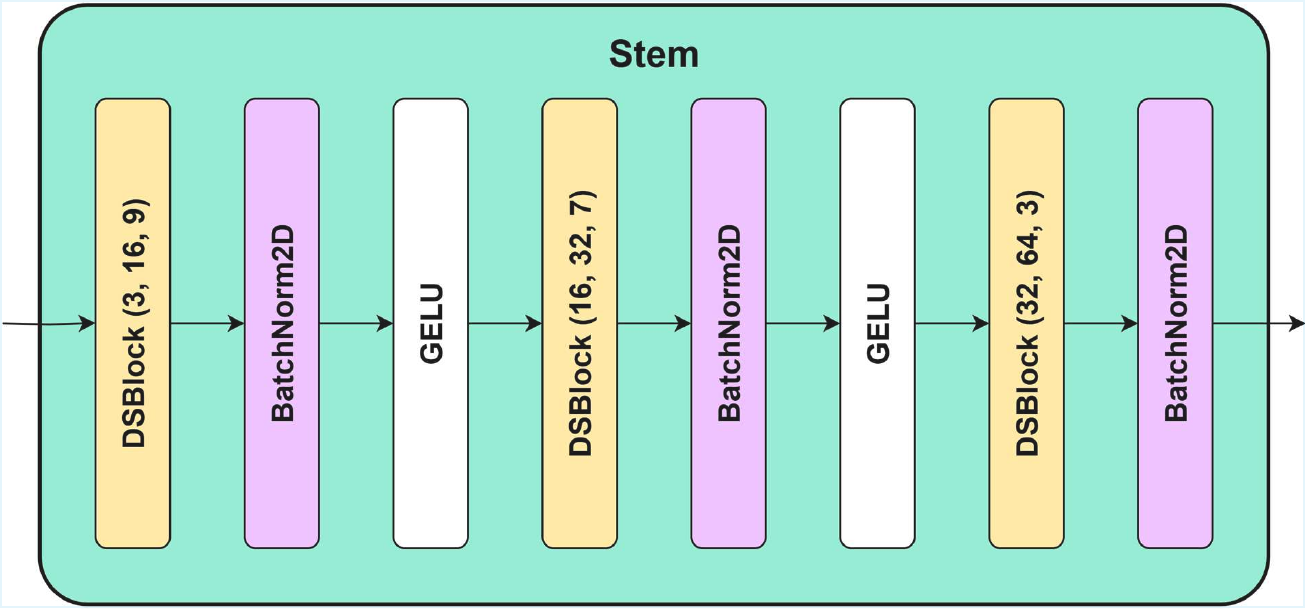}
\caption{Stem architecture of OpenVeinNet, featuring a sequence of three Dynamic Snake Convolution (DSConv) layers with kernel sizes \(K \in \{9, 7, 3\}\), stride \(2\), and GELU activation, processing an input image \(\bm{I} \in \mathbb{R}^{B \times 3 \times 224 \times 224}\) to produce a feature map \(\bm{X}_{\text{stem}} \in \mathbb{R}^{B \times 64 \times 28 \times 28}\) for the backbone.}
\label{fig:stem}
\end{figure}

Figure~\ref{fig:stem} shows the complete stem architecture. Each DSConv Block is followed by batch normalisation and GELU activation. The GELU activation is omitted after the final DSConv Block to allow the extracted feature map to retain a wider range of responses, which is useful for preserving fine-grained vein information before graph-based processing.

Given a finger vein image $\bm{X} \in \mathbb{R}^{B \times C_{\text{in}} \times H \times W}$, where $B$ is the batch size, $C_{\text{in}}$ is the number of channels, and $H = W = 224$, each DSConv applies a deformable sampling strategy based on learned offset maps. A \(3 \times 3\) convolution predicts offsets \(\bm{O} \in \mathbb{R}^{B \times 2K \times H \times W}\), where \(K\) is the kernel size and \(\bm{O} = (\bm{O}_y, \bm{O}_x)\) contains offsets for the vertical and horizontal directions:
\begin{equation}
    \bm{O} = \tanh(\mathrm{GroupNorm}(\mathrm{Conv}_{3 \times 3}(\bm{X})))
\end{equation}

Let \(\bm{p} = (p_y, p_x)\) denote a pixel position in the spatial grid, where \(p_y \in \{0,1,\dots,H-1\}\) and \(p_x \in \{0,1,\dots,W-1\}\). The coordinate maps are generated as:
\begin{equation}
    \bm{Y}_{\text{coord}}, \bm{X}_{\text{coord}} = \mathrm{GetCoordinateMap2D}(\bm{O}, m, \mathrm{extend\_scope})
\end{equation}
where \(m \in \{0,1\}\) specifies deformation along the \(x\)-axis (\(m=0\)) or \(y\)-axis (\(m=1\)), and \(\mathrm{extend\_scope}=1.0\) controls the deformation range. The coordinates are normalised to \([-1,1]\) before interpolation.

The deformed feature map is sampled using bilinear interpolation:
\begin{equation}
    \bm{X}_{\text{deformed}} = \mathrm{GridSample}(\bm{X}, [\bm{X}_{\text{coord}}, \bm{Y}_{\text{coord}}], \text{mode}=\text{"bilinear"})
\end{equation}

A directional convolution is then applied to the deformed feature map:
\begin{equation}
    \bm{X}_{\text{dir}} = 
    \begin{cases} 
    \mathrm{Conv}_{(K, 1)}(\bm{X}_{\text{deformed}}), & \text{if } m = 0 \\
    \mathrm{Conv}_{(1, K)}(\bm{X}_{\text{deformed}}), & \text{if } m = 1
    \end{cases}
\end{equation}

The output is normalised and activated:
\begin{equation}
    \bm{X}_{\text{dir}} = \mathrm{ReLU}(\mathrm{GroupNorm}(\bm{X}_{\text{dir}})).
\end{equation}

The outputs from the standard convolution branch \(\bm{X}_{\text{conv}} = \mathrm{Conv}_{K \times K}(\bm{X})\), the \(x\)-axis DSConv branch \(\bm{X}_x = \text{DSConv}_x(\bm{X})\), and the \(y\)-axis DSConv branch \(\bm{X}_y = \text{DSConv}_y(\bm{X})\) are concatenated and fused by a final convolution:
\begin{equation}
    \bm{X}_{\text{out}} = \mathrm{Conv}_{K \times K}(\mathrm{Concat}(\bm{X}_{\text{conv}}, \bm{X}_x, \bm{X}_y)).
\end{equation}

The adaptive offset learning in DSConv can be interpreted as aligning the sampling locations with vein-like structures:
\begin{equation}
\min_{\bm{O}} \sum_{\text{pixels}} \|\bm{X}(\bm{p} + \bm{O}) - \bm{V}(\bm{p})\|_2^2,
\end{equation}
where \(\bm{V}(\bm{p})\) denotes an idealised vein response at location \(\bm{p}\). Although \(\bm{V}(\bm{p})\) is not explicitly supervised, this formulation illustrates the intended behaviour of the learned offsets: to adapt the receptive field towards vein-aligned structures.

The final output of the stem is written as:
\begin{equation}
    \bm{X}_{\text{stem}} = \mathrm{Stem}(\bm{I}) \in \mathbb{R}^{B \times C_{\text{out}} \times \frac{H}{2^d} \times \frac{W}{2^d}},
\end{equation}
where \(C_{\text{out}} = 64\), \(H=W=224\), and \(d=3\), resulting in \(\bm{X}_{\text{stem}} \in \mathbb{R}^{B \times 64 \times 28 \times 28}\).

To preserve spatial information, a learnable positional embedding \(\bm{P} \in \mathbb{R}^{1 \times C_{\text{out}} \times \frac{H}{2^d} \times \frac{W}{2^d}}\) is added to the stem output:
\begin{equation}
    \bm{X}_{\text{stem}} = \mathrm{Stem}(\bm{I}) + \bm{P}.
\end{equation}
The positional embedding provides spatial context before graph construction, helping the backbone retain the topological arrangement of vein structures.

\subsection{Backbone: Grapher Blocks}
\label{backbone}

The backbone of \textit{OpenVeinNet} is designed to model long-range connectivity between vein regions using Graph Convolutional Networks (GCNs). The stem output, augmented with positional embeddings, is processed by a sequence of Grapher Blocks. Each Grapher Block constructs a k-nearest-neighbour (k-NN) graph over feature nodes and aggregates information from neighbouring nodes. This enables the model to capture non-local relationships that are difficult to represent using only standard convolutional operations.

\begin{figure}[!ht] 
\centering
\includegraphics[width=0.325\textwidth]{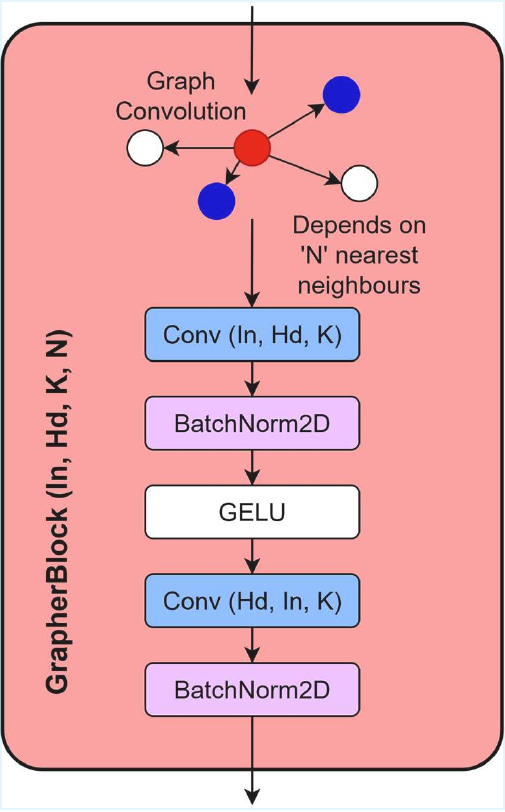}
\caption{GrapherBlock (In: input channels, Out: output channels, K: kernel size, N: nearest neighbours taken into account by graph convolution.)}
\label{fig:grapherblock}
\end{figure}

Given the input \(\bm{X}_{\text{stem}} \in \mathbb{R}^{B \times 64 \times 28 \times 28}\), each Grapher Block operates as follows:
\begin{itemize}
    \item \textbf{Initial Projection}: A \(1 \times 1\) convolution refines the input features, followed by batch normalisation:
    \begin{equation}
        \bm{Y} = \mathrm{BatchNorm}(\mathrm{Conv}_{1 \times 1}(\bm{X}_{\text{stem}})).
    \end{equation}

    \item \textbf{Dynamic Graph Construction}: A k-NN graph is constructed based on feature similarity:
    \begin{equation}
    \begin{aligned}
    \mathrm{EdgeIndex} = \mathrm{DenseDilatedKnnGraph}(&\bm{Y},\ k,\ \text{dilation}, \\
                                                  &\text{stochastic},\ \epsilon).
    \end{aligned}
    \end{equation}
    We use \(k=18\) in the first stage and \(k=9\) in the second stage.

    \item \textbf{Graph Convolution}: Features from neighbouring nodes are aggregated using an MRConv operation:
    \begin{equation}
        \bm{X}_{\text{out}} = \mathrm{Max}(\mathrm{BasicConv}(\mathrm{Concat}(\bm{Y}_i, \bm{Y}_j - \bm{Y}_i))),
    \end{equation}
    where \(\bm{Y}_i\) and \(\bm{Y}_j\) denote the features of node \(i\) and its neighbour \(j\), respectively.

    \item \textbf{Feed-Forward and Residual Connection}: A feed-forward network with two \(1 \times 1\) convolutions refines the graph-aggregated features, and a residual connection is used to preserve the original feature information.
\end{itemize}

The graph convolution can be interpreted as aggregating the strongest relational evidence among the neighbouring nodes:
\begin{equation}
    \bm{X}_{\text{out}} \approx \max_{j \in \mathcal{N}(i)} \mathrm{Sim}(\bm{Y}_i, \bm{Y}_j),
\end{equation}
where \(\mathcal{N}(i)\) denotes the set of \(k\) nearest neighbours of node \(i\).

\begin{figure}[!ht]
\centering
\includegraphics[width=0.425\textwidth]{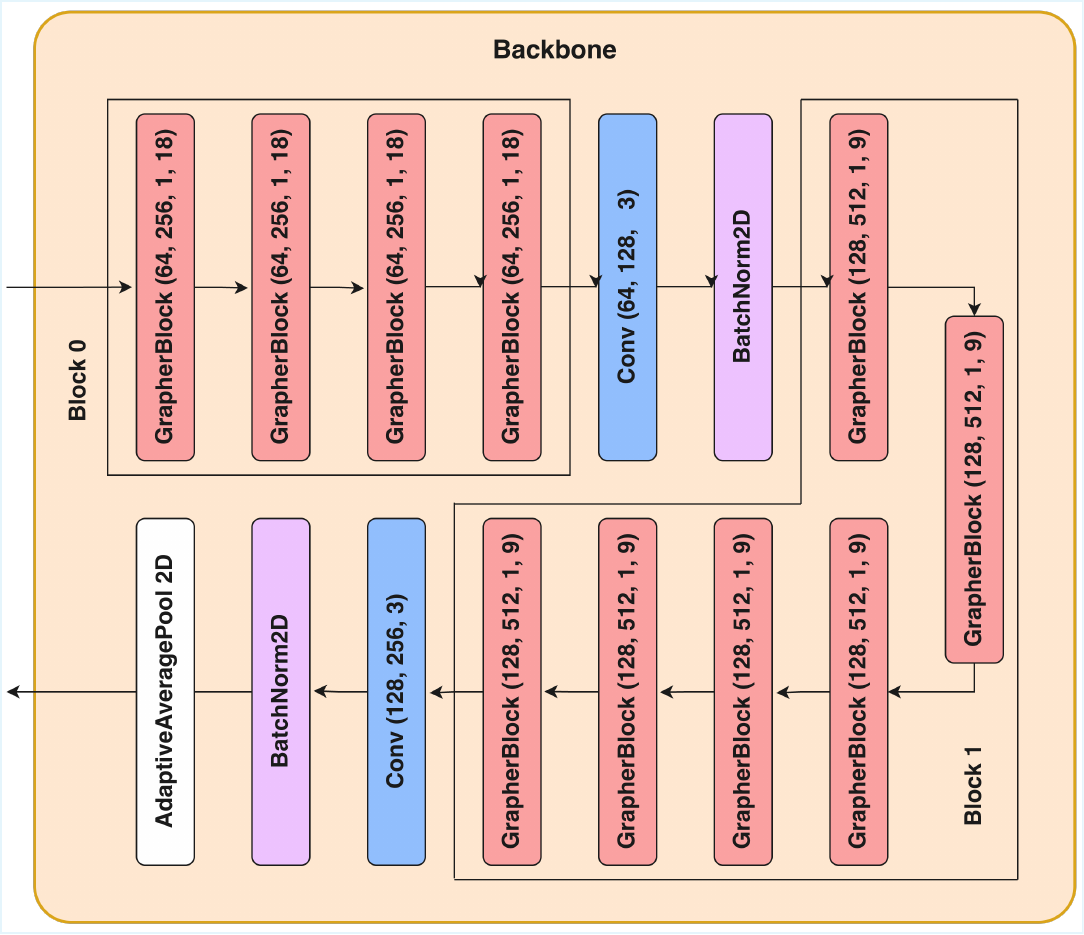}
\caption{Backbone structure of OpenVeinNet, with four sequential Grapher Blocks, a convolution layer with batch normalisation, six additional Grapher Blocks, another convolution layer with batch normalisation, and adaptive average pooling.}
\label{fig:backbone}
\end{figure}

As shown in Figure~\ref{fig:backbone}, the backbone has two graph-processing stages followed by pooling. \textbf{Block 0} takes \(64\) input channels and produces \(128\) output channels, using \(k=18\) neighbours and a \(3 \times 3\) shrinker convolution with stride \(2\), reducing the spatial resolution to \(14 \times 14\). This stage contains four internal Grapher units. \textbf{Block 1} takes \(128\) input channels and produces \(256\) output channels, using \(k=9\) neighbours and a \(3 \times 3\) shrinker convolution with stride \(2\), reducing the spatial resolution to \(7 \times 7\). This stage contains six internal Grapher units.

Finally, adaptive average pooling produces the embedding vector:
\begin{equation}
    \bm{E} = \mathrm{AdaptiveAvgPool2d}(\mathrm{GrapherBlocks}(\bm{X}_{\text{stem}})) \in \mathbb{R}^{B \times 256}.
\end{equation}
The resulting embedding \(\bm{E}\) is used for cosine-similarity-based verification. The use of k-NN graph construction enables the model to capture topological relationships between vein regions, while the pooled embedding provides a compact representation suitable for open-set verification.

\subsection{Centroid Angular Hybrid Loss}
\label{sec:loss}

The softmax loss is commonly used for classification tasks and is defined as:
\begin{equation}
\mathcal{L}_1 = -\log\frac{e^{W_{y_i}^T x_i + b_{y_i}}}{\sum_{j=1}^{N} e^{W_j^T x_i + b_j}},
\label{eq:l1}
\end{equation}
where \(x_i \in \mathbb{R}^d\) represents the embedding of the \(i\)-th sample with class label \(y_i\), \(W_j \in \mathbb{R}^d\) denotes the \(j\)-th class weight vector from \(W \in \mathbb{R}^{d \times N}\), \(b_j\) is the bias term, and \(N\) is the number of training classes. 

Although softmax loss is effective for closed-set classification, it does not explicitly enforce compact intra-class embeddings or angularly separated inter-class representations. This limitation is important in open-set verification, where the model must generalise beyond the identities seen during training and reject non-enrolled probes using a similarity threshold. To address this, we introduce the \textit{Centroid Angular Hybrid Loss} \((\mathcal{L}_2)\), which augments the softmax objective with an angular term between the sample embedding and an independently learned class-direction reference.

The proposed loss is defined as:
\begin{equation}
\mathcal{L}_2 = -\log\frac{e^{W_{y_i}^T x_i + b_{y_i} + \beta \cdot \theta_{y_i}}}{\left( \sum_{j=1}^{N} e^{W_j^T x_i + b_j} \right) \cdot \left( \sum_{j=1}^{N} e^{\theta_j} \right)}.
\label{eq:l2}
\end{equation}

The angular component \(\theta_j\) is defined as:
\begin{equation}
\theta_j = \cos^{-1} \left( \frac{\tilde{W}_j^T x_i}{\|\tilde{W}_j\| \cdot \|x_i\|} \right),
\end{equation}
where \(\tilde{W} \in \mathbb{R}^{d \times N}\) has the same shape as \(W\), but is initialised independently. The term \(\theta_j\) captures the angular misalignment between the embedding \(x_i\) and the centroid-like direction \(\tilde{W}_j\).

The balancing factor \(\beta\) controls the contribution of the angular component. A small value of \(\beta\) weakens angular supervision, while an overly large value can dominate the optimisation and reduce stability. The proposed formulation therefore encourages correct class prediction while also shaping the embedding space to be more compact within identities and more separated across identities. This is beneficial for open-set verification because the final decision is based on cosine similarity between enrolment and probe embeddings. In this formulation, the term centroid refers to a learnable class-direction prototype in the embedding space rather than an empirical mini-batch centroid.

In summary, the Centroid Angular Hybrid Loss complements the DSConv-GCN architecture by improving the discriminability of the learned feature space. DSConv extracts local curvilinear vein evidence, the graph backbone models long-range vascular relationships, and the proposed loss encourages embeddings that are better suited for threshold-based verification of enrolled and non-enrolled identities.
\section{Experiments and Results}
\label{sec:experiments}

This section presents the experimental evaluation of the proposed \textit{OpenVeinNet} and compares it with existing finger vein verification methods. Performance is reported using the Area Under the Curve (AUC), Equal Error Rate (EER), and threshold-dependent operating points, namely TAR@FAR$=0.1\%$, TAR@FAR$=1\%$, and TAR@FAR$=10\%$. EER is computed at the operating point where the False Match Rate (FMR) equals the False Non-Match Rate (FNMR). The proposed method is compared with handcrafted and deep learning-based finger vein verification methods, including MCP~\cite{surveyonmanualfeatures}, RLT~\cite{surveyonmanualfeatures}, WLD~\cite{surveyonmanualfeatures}, multiple attention~\cite{binma}, deep fusion~\cite{ism}, self-attention~\cite{veinatnnet}, ECA-ResNet with arc-cosine loss~\cite{arcvein}, vision transformers~\cite{vit}, and the local-global feature interaction network~\cite{lgfin}. Training details for all baselines are provided in the supplementary material. The experiments are conducted on five publicly available finger vein datasets to evaluate both cross-dataset generalisation and open-set verification behaviour.

\subsection{Datasets}

The experiments utilise five publicly available finger vein datasets: FV-300 (A)~\cite{veinatnnet}, MMCBNU-6000 (B)~\cite{mmcbnu}, FV-USM (C)~\cite{fvusm}, PolyU (D)~\cite{polyu}, and VERA (E)~\cite{vera}. A summary of these datasets is provided in Table~\ref{tab:datasets}. A brief description of each dataset is given below.

\begin{itemize}
    \item \textbf{FV-300}~\cite{veinatnnet}: This dataset contains 300 distinct finger vein patterns collected from 75 subjects. Images were captured using a custom monochromatic CMOS camera with two lighting sources. For each subject, the index and middle fingers of both hands were imaged across multiple sessions, with intervals ranging from 1 to 4 days. Among the datasets used in this study, FV-300 provides the highest image quality. For our experiments, a total of 124 images were removed during preprocessing due to severe blur, poor illumination, or failed vein visibility, as these conditions prevent reliable extraction of vein structures.

    \item \textbf{MMCBNU-6000}~\cite{mmcbnu}: Developed by Chonbuk National University, this dataset includes 600 distinct finger vein patterns from 100 individuals. Acquisition was performed using a camera equipped with an NIR-pass filter and an array of 850~nm infrared LEDs. For each subject, the index, middle, and ring fingers of both hands were captured ten times. The dataset includes participants of diverse nationalities, age groups, blood types, and genders.

    \item \textbf{FV-USM}~\cite{fvusm}: Collected at Universiti Sains Malaysia, this dataset comprises 492 distinct finger vein patterns from 123 subjects, including 83 males and 40 females aged between 20 and 52. Each subject contributed images of the index and middle fingers of both hands. Data were acquired in two sessions separated by a two-week interval.

    \item \textbf{PolyU}~\cite{polyu}: This dataset contains 312 distinct finger vein patterns from 156 subjects, captured using a peg-free and unconstrained imaging setup. Data acquisition spanned an 11-month period. In this study, we use version~1 of the dataset, which consists of 218 unique finger vein patterns, with four images per finger.

    \item \textbf{VERA}~\cite{vera}: This dataset was acquired using an open finger vein sensor and contains 220 distinct finger vein patterns from 110 subjects. Each subject presented the index fingers of both hands, with two samples captured per finger at a 5-minute interval. The dataset includes 40 female and 70 male participants aged between 18 and 60, with a mean age of 33 years.
\end{itemize}

\begin{table}[hbt!]
    \centering
    \begin{tabular}{|c|c|c|c|}
         \hline
         Dataset & Classes & Images per identity & Total Images \\
         \hline
         \hline
         FV-300 (A)~\cite{veinatnnet} & 300 & 92* & 27,600* \\
         \hline
         MMCBNU (B)~\cite{mmcbnu} & 600 & 10 & 6,000 \\
         \hline
         FV-USM (C)~\cite{fvusm} & 492 & 12 & 5,904 \\
         \hline
         PolyU (D)~\cite{polyu} & 218 & 4 & 872 \\
         \hline
         VERA (E)~\cite{vera} & 220 & 2 & 440 \\
         \hline
    \end{tabular}
    \medskip
    \caption{Summary of the five public finger vein datasets used in this study. *A total of 124 low-quality FV-300 images were removed due to severe blur, poor illumination, or failed vein visibility.*}
    \label{tab:datasets}
\end{table}

\subsection{Evaluation Protocol}
\label{sec:evaluation_protocol}

We evaluate \textit{OpenVeinNet} using a leave-one-dataset-out strategy. In each experiment, the model is trained on four datasets and evaluated on the remaining held-out dataset. Therefore, all identities in the evaluation dataset are unseen during training. This setting allows us to assess both cross-dataset generalisation and open-set verification behaviour under sensor, subject, and acquisition-domain shifts.

\begin{table}[ht!]
    \centering
    \scriptsize
    \begin{tabular}{|c|c|c|c|c|}
        \hline
        \multirow{2}{*}{Dataset} & \multicolumn{2}{c|}{Genuine Scores} & \multicolumn{2}{c|}{Impostor Scores} \\
        \cline{2-5}
         & Half & Full & Half & Full \\
        \hline
        FV-300 (A)~\cite{veinatnnet} & 598,279 & 1,198,364 & 181,709,824 & 362,208,956 \\
        \hline
        MMCBNU (B)~\cite{mmcbnu} & 13,500 & 27,000 & 9,000,000 & 17,970,000 \\
        \hline
        FV-USM (C)~\cite{fvusm} & 16,236 & 32,472 & 8,714,304 & 17,393,184 \\
        \hline
        PolyU (D)~\cite{polyu} & 1,562 & 3,107 & 405,765 & 807,794 \\
        \hline
        VERA (E)~\cite{vera} & 110 & 220 & 48,400 & 96,360 \\
        \hline
    \end{tabular}
    \medskip
    \caption{Number of genuine and impostor scores used for the half-subject and full-subject verification protocols.}
    \label{tab:rebuttal_score_counts}
\end{table}

We consider two complementary evaluation protocols.

\textbf{Protocol 1: Enrollment-based unknown-rejection evaluation.}
The held-out evaluation dataset is deterministically split into two disjoint subject groups. The first 50\% of subjects are treated as the enrolled set, while the remaining 50\% are treated as non-enrolled subjects. Genuine scores are computed from all within-subject pairwise cosine similarities among the enrolled subjects only. Impostor scores are computed from all cross-subject pairwise cosine similarities between enrolled-subject samples and non-enrolled-subject samples. Threshold-dependent metrics are then obtained by sweeping the decision threshold over these score distributions.

\textbf{Protocol 2: Full-subject verification evaluation.}
In the implemented full-subject protocol, all held-out subjects are included in evaluation and genuine/impostor scores are computed from all possible within-subject and cross-subject pairwise cosine similarities over the held-out set. Thus, the reported FAR/FRR, EER, AUC, and TAR values are derived from exhaustive pairwise score distributions rather than from a separate gallery-template matching stage.

For both protocols, genuine and impostor scores are computed using cosine similarity between the extracted embeddings.  Genuine scores are generated by comparing samples belonging to the same finger identity, excluding self-comparisons. Impostor scores are generated by comparing samples from different finger identities. Unless otherwise stated, all possible genuine and impostor pairs are used for evaluation without random pair sampling.
Performance is reported using AUC, EER, and TAR at fixed FAR values. The threshold-based operating points, TAR@FAR$=0.1\%$, TAR@FAR$=1\%$, and TAR@FAR$=10\%$, are included because practical biometric systems are deployed at fixed decision thresholds rather than only at the equal-error operating point.

\textbf{Cross-validation and confidence intervals.}
To improve statistical reliability, we repeat each leave-one-dataset-out experiment over multiple predefined statistical splits indexed by \texttt{stat\_seed}. For each \texttt{stat\_seed}, the corresponding dataset partition is loaded from \texttt{data/<dataset>/<stat\_seed>}, and all methods are evaluated on the same held-out dataset using the same score-generation protocol. The final results are reported as the mean and 95\% confidence interval (CI) across the available matched runs for each protocol.

\subsection{Results and Discussion}
\label{sec:results_discussion}
Tables~\ref{tab:rebuttal_half_thresholds} and~\ref{tab:rebuttal_full_thresholds} present the results under the enrollment-based unknown-rejection and full-subject verification protocols, respectively. The results are reported across five leave-one-dataset-out settings. Overall, the proposed method achieves strong and stable performance across both protocols, demonstrating that the learned embeddings generalise well to unseen identities and different acquisition conditions.

\textbf{Enrollment-based unknown-rejection performance.}
Table~\ref{tab:rebuttal_half_thresholds} reports the stricter open-set protocol, where only half of the subjects from the held-out dataset are enrolled and the remaining subjects are treated as unknown identities. Under this setting, the proposed method achieves the best or highly competitive performance across most evaluation splits. In the challenging \(ABDE \rightarrow C\) setting, OpenVeinNet obtains an AUC of \(97.72\%\) and an EER of \(7.48\%\), clearly outperforming the deep learning baselines. At the operating-point level, it achieves \(53.52\%\) TAR at FAR$=1\%$ and \(95.47\%\) TAR at FAR$=10\%\), showing that the learned embeddings support reliable acceptance of enrolled users while maintaining rejection capability for unknown probes.

A similar trend is observed for \(ABCE \rightarrow D\), where the proposed method achieves the best AUC (\(98.67\%\)), lowest EER (\(5.34\%\)), and the strongest TAR values across the reported operating points. These results are important because they directly address the unknown-rejection setting: the method is not only evaluated on unseen subjects but also tested for its ability to reject non-enrolled identities at fixed thresholds.

\begin{table*}[ht!]
    \centering
    \scriptsize
    \resizebox{\textwidth}{!}{
    \begin{tabular}{|c|c|c|c|c|c|c|}
        \hline
        $Train \rightarrow Test$ Dataset & Algorithm & AUC (\%) & EER (\%) & TAR@FAR$=0.1\%$ (\%) & TAR@FAR$=1\%$ (\%) & TAR@FAR$=10\%$ (\%) \\
        \hline
        $BCDE \rightarrow A$ & MCP \cite{mcp} & 99.80 & 0.89 & 40.19 & 99.47 & \textbf{100.00} \\
         & RLT \cite{rlt} & 99.78 & 0.95 & 30.00 & 99.24 & \textbf{100.00} \\
         & WLD \cite{wld} & \textbf{99.82} & \textbf{0.84} & 43.82 & \textbf{99.59} & 100.00 \\
         & ArcVein \cite{arcvein} & 89.39 (82.64--96.14) & 19.49 (11.45--27.54) & 6.67 (0.00--16.88) & 22.81 (2.86--42.77) & 61.22 (35.63--86.81) \\
         & LGFIN \cite{lgfin} & 85.52 (83.95--87.08) & 23.35 (22.54--24.15) & 4.23 (0.00--8.51) & 15.12 (6.50--23.74) & 50.09 (41.24--58.95) \\
         & FV-ViT \cite{vit} & 89.72 (88.31--91.13) & 17.74 (16.16--19.33) & 0.95 (0.00--2.12) & 6.34 (3.56--9.12) & 58.68 (51.23--66.13) \\
         & VeinAttNet \cite{veinatnnet} & 97.83 (95.75--99.91) & 6.56 (2.64--10.48) & 21.55 (0.00--54.93) & 50.87 (12.87--88.86) & 96.75 (91.84--100.00) \\
         & Proposed Method & 99.47 (99.19--99.75) & 3.49 (2.47--4.51) & \textbf{58.11} (42.09--74.13) & 85.03 (76.76--93.30) & 99.80 (99.53--100.00) \\
        \hline
        $ACDE \rightarrow B$ & MCP \cite{mcp} & 92.90 & 12.31 & 0.04 & 3.29 & 77.65 \\
         & RLT \cite{rlt} & 90.31 & 13.76 & 0.00 & 0.04 & 57.20 \\
         & WLD \cite{wld} & 87.76 & 16.98 & 0.00 & 0.06 & 37.44 \\
         & ArcVein \cite{arcvein} & 91.71 (86.06--97.37) & 15.61 (8.81--22.41) & 0.99 (0.00--2.30) & 18.34 (5.65--31.02) & 72.04 (48.42--95.66) \\
         & LGFIN \cite{lgfin} & 92.48 (91.94--93.02) & 15.34 (14.66--16.02) & 5.26 (1.47--9.06) & 26.98 (25.83--28.13) & 75.30 (72.60--77.99) \\
         & FV-ViT \cite{vit} & 89.70 (87.28--92.12) & 18.03 (15.65--20.41) & 1.94 (0.82--3.06) & 14.97 (7.27--22.68) & 64.48 (54.60--74.36) \\
         & VeinAttNet \cite{veinatnnet} & 99.06 (98.74--99.39) & 4.35 (3.82--4.88) & \textbf{39.09} (27.25--50.92) & \textbf{78.50} (69.77--87.22) & 98.79 (98.37--99.21) \\
         & Proposed Method & \textbf{99.12} (98.86--99.38) & \textbf{3.99} (3.29--4.70) & 28.13 (19.96--36.30) & 74.71 (67.50--81.91) & \textbf{99.51} (99.18--99.84) \\
        \hline
        $ABDE \rightarrow C$ & MCP \cite{mcp} & 87.01 & 18.93 & 0.01 & 0.55 & 35.11 \\
         & RLT \cite{rlt} & 79.40 & 25.14 & 0.00 & 0.00 & 4.91 \\
         & WLD \cite{wld} & 83.07 & 22.87 & 0.00 & 0.08 & 20.58 \\
         & ArcVein \cite{arcvein} & 87.51 (84.43--90.58) & 20.22 (16.79--23.64) & 2.02 (0.00--4.36) & 7.84 (3.90--11.78) & 52.75 (41.61--63.90) \\
         & LGFIN \cite{lgfin} & 87.78 (86.39--89.17) & 18.34 (14.20--22.47) & 0.81 (0.00--2.13) & 4.72 (0.00--10.76) & 47.65 (15.14--80.17) \\
         & FV-ViT \cite{vit} & 89.66 (88.43--90.89) & 18.24 (17.19--19.28) & 2.23 (0.95--3.51) & 15.78 (8.26--23.30) & 65.69 (61.22--70.17) \\
         & VeinAttNet \cite{veinatnnet} & 86.79 (85.37--88.22) & 22.20 (20.14--24.26) & 8.01 (5.92--10.10) & 15.23 (11.25--19.21) & 49.21 (46.92--51.50) \\
         & Proposed Method & \textbf{97.72} (97.28--98.17) & \textbf{7.48} (6.70--8.25) & \textbf{13.71} (3.69--23.72) & \textbf{53.52} (47.25--59.79) & \textbf{95.47} (93.72--97.21) \\
        \hline
        $ABCE \rightarrow D$ & MCP \cite{mcp} & 95.77 & 9.90 & 3.83 & 21.88 & 90.35 \\
         & RLT \cite{rlt} & 88.17 & 16.51 & 0.00 & 0.07 & 34.81 \\
         & WLD \cite{wld} & 89.72 & 15.60 & 0.00 & 0.86 & 48.95 \\
         & ArcVein \cite{arcvein} & 83.78 (82.07--85.48) & 24.18 (22.21--26.14) & 1.43 (0.00--5.53) & 5.44 (1.23--9.66) & 46.83 (38.51--55.15) \\
         & LGFIN \cite{lgfin} & 93.95 (92.23--95.66) & 13.41 (11.36--15.45) & 15.41 (10.38--20.43) & 32.25 (18.67--45.83) & 79.50 (71.57--87.43) \\
         & FV-ViT \cite{vit} & 87.65 (85.36--89.93) & 20.32 (18.21--22.42) & 0.67 (0.00--1.73) & 11.18 (2.88--19.48) & 59.39 (52.00--66.78) \\
         & VeinAttNet \cite{veinatnnet} & 97.90 (97.34--98.47) & 7.72 (6.71--8.72) & 24.96 (0.00--52.09) & 67.50 (58.20--76.79) & 94.53 (93.02--96.05) \\
         & Proposed Method & \textbf{98.67} (98.32--99.02) & \textbf{5.34} (4.40--6.29) & \textbf{25.65} (11.73--39.56) & \textbf{67.68} (60.38--74.98) & \textbf{98.39} (97.81--98.97) \\
        \hline
        $ABCD \rightarrow E$ & MCP \cite{mcp} & 72.21 & 33.19 & \textbf{0.00} & 0.04 & 11.14 \\
         & RLT \cite{rlt} & 59.21 & 44.43 & \textbf{0.00} & 0.00 & 1.11 \\
         & WLD \cite{wld} & 69.67 & 35.45 & \textbf{0.00} & 0.02 & 8.51 \\
         & ArcVein \cite{arcvein} & 71.51 (69.25--73.77) & 33.66 (31.29--36.04) & \textbf{0.00} (0.00--0.00) & 0.38 (0.05--0.71) & 18.16 (15.53--20.79) \\
         & LGFIN \cite{lgfin} & 75.88 (74.57--77.20) & 32.29 (31.43--33.14) & \textbf{0.00} (0.00--0.00) & 9.23 (0.00--20.97) & 32.59 (23.47--41.70) \\
         & FV-ViT \cite{vit} & 74.34 (72.58--76.09) & 32.09 (28.61--35.57) & \textbf{0.00} (0.00--0.00) & 4.10 (0.00--8.95) & 30.29 (23.60--36.98) \\
         & VeinAttNet \cite{veinatnnet} & 89.73 (88.62--90.83) & 18.19 (17.04--19.35) & \textbf{0.00} (0.00--0.00) & 13.49 (0.63--26.36) & \textbf{67.34} (56.41--78.27) \\
         & Proposed Method & \textbf{91.32} (87.69--94.95) & \textbf{16.64} (13.57--19.71) & \textbf{0.00} (0.00--0.00) & \textbf{13.70} (0.00--31.29) & 66.40 (45.03--87.76) \\
        \hline
    \end{tabular}}
    \caption{Half-subject threshold-based operating-point results.}
    \label{tab:rebuttal_half_thresholds}
\end{table*}

\textbf{Full-subject verification performance.}
Table~\ref{tab:rebuttal_full_thresholds} reports the results when all subjects from the held-out dataset are enrolled. In this setting, the proposed method achieves the lowest EER in 4 of 5 evaluation splits and is competitive in the 5th one. For example, in the \(BCDE \rightarrow A\) setting, OpenVeinNet achieves an EER of \(3.17\%\), compared with \(6.20\%\) for VeinAttNet and substantially higher EER values for other deep learning baselines. In the more difficult \(ABDE \rightarrow C\) and \(ABCE \rightarrow D\) settings, the proposed method obtains EER values of \(7.44\%\) and \(5.21\%\), respectively, while also achieving strong TAR values at low FAR thresholds. These results show that the embedding space learned by OpenVeinNet remains discriminative under substantial dataset shift.

\textbf{Behaviour at strict operating points.}
From a deployment perspective, the low-FAR regime is particularly important because false accepts directly affect security. The proposed method generally achieves higher TAR values at FAR$=0.1\%$ and FAR$=1\%$ than the deep learning baselines, especially in the challenging \(ABDE \rightarrow C\) and \(ABCE \rightarrow D\) protocols. Handcrafted methods such as MCP and WLD can perform strongly on clean and well-controlled data, particularly in \(BCDE \rightarrow A\), but their performance degrades considerably under stronger domain shift. This suggests that handcrafted vessel-enhancement pipelines can be effective when the target data are clean, but their robustness is limited when image quality, sensor characteristics, and acquisition conditions vary.

One exception to the above trend is observed in the \textit{ACDE}$\rightarrow$\textit{B} 
protocol at FAR$=$0.1\%, where VeinAttNet achieves a higher TAR of 39.09\% compared  with 28.13\% for the proposed method. This can be explained by the score distribution  behaviour on MMCBNU, which has the highest gradient coherence score among all five  datasets (0.5398, Table~7), reflecting strongly defined and structurally consistent vein  patterns. Under such controlled acquisition conditions, VeinAttNet's attention-based 
architecture produces genuine scores that are more tightly concentrated near the upper  end of the cosine similarity range, which is advantageous specifically at the strict  FAR$=$0.1\% threshold. The graph-based embeddings of OpenVeinNet produce a smoother score distribution that is better calibrated across a wider range of operating points, 
as evidenced by the lower EER (3.99\% vs.\ 4.35\%) and competitive TAR at FAR$=$1\% and FAR$=$10\%. This trade-off between strict threshold performance and overall operating-point calibration is consistent with the dataset quality analysis in Section~\ref{sec:dataset_quality_analysis}.

\textbf{Challenging dataset analysis.}
The \(ABCD \rightarrow E\) setting is the most challenging protocol because VERA contains only two images per finger identity/class and exhibits higher acquisition variability. All methods show reduced performance in this setting. Nevertheless, the proposed method achieves the best AUC and EER in both evaluation protocols. At FAR$=0.1\%$, the TAR is zero for all methods, indicating that extremely strict operating points are difficult to realise reliably for this low-sample dataset. This should be interpreted with care: the small number of genuine comparisons in VERA leads to a coarse score distribution, which limits score resolution at very low FAR thresholds. Hence, the zero TAR at FAR$=0.1\%$ reflects the limited score resolution and strict operating threshold rather than a complete failure of the learned representation. Importantly, OpenVeinNet maintains competitive or best performance at FAR$=1\%$ and FAR$=10\%$, demonstrating better relative robustness under severe domain shift.

\textbf{Comparison with state-of-the-art methods.}
Compared with recent deep learning approaches such as ArcVein, LGFIN, FV-ViT, and VeinAttNet, the proposed method provides a stronger overall trade-off between AUC, EER, and threshold-based TAR. While VeinAttNet is competitive in some settings, such as \(ACDE \rightarrow B\), its performance is less stable across all held-out datasets. In contrast, OpenVeinNet achieves consistently low EER and strong TAR across both protocols, demonstrating more reliable generalisation.

\textbf{Technical justification of performance gains.}
The improvement of OpenVeinNet can be attributed to the complementarity of its three main components. First, DSConv adaptively samples along curvilinear structures and therefore captures local tubular vein evidence more effectively than fixed-grid convolution. Second, the graph-based backbone models long-range relationships between vein regions, enabling the network to encode global vascular topology. Third, the Centroid Angular Hybrid Loss encourages compact intra-class embeddings and angularly separated inter-class representations, which improves cosine-similarity-based verification under open-set conditions.

Overall, the experimental results show that OpenVeinNet improves cross-dataset generalisation and provides strong open-set verification capability. The consistent gains in EER and threshold-based TAR across several held-out datasets support the effectiveness of combining adaptive tubular feature extraction, graph-based relational modelling, and angularly constrained embedding learning.
\begin{table*}[ht!]
    \centering
    \scriptsize
    \resizebox{\textwidth}{!}{
        \begin{tabular}{|c|c|c|c|c|c|c|}
        \hline
        $Train \rightarrow Test$ Dataset & Algorithm & AUC (\%) & EER (\%) & TAR@FAR$=0.1\%$ (\%) & TAR@FAR$=1\%$ (\%) & TAR@FAR$=10\%$ (\%) \\
        \hline
        $BCDE \rightarrow A$ & MCP \cite{mcp} & 99.80 & 0.89 & 40.19 & 99.47 & \textbf{100.00} \\
         & RLT \cite{rlt} & 99.78 & 0.95 & 30.00 & 99.24 & \textbf{100.00} \\
         & WLD \cite{wld} & \textbf{99.82} & \textbf{0.84} & 43.82 & \textbf{99.59} & 100.00 \\
         & ArcVein \cite{arcvein} & 89.77 (83.13--96.42) & 19.38 (10.81--27.95) & 6.63 (1.15--12.10) & 24.80 (5.88--43.72) & 63.72 (39.50--87.95) \\
         & LGFIN \cite{lgfin} & 85.65 (84.32--86.99) & 23.01 (22.30--23.72) & 3.28 (1.06--5.50) & 14.68 (6.02--23.35) & 50.09 (42.60--57.59) \\
         & FV-ViT \cite{vit} & 89.87 (88.51--91.23) & 17.37 (15.84--18.89) & 1.04 (0.00--2.11) & 6.70 (3.57--9.84) & 59.09 (51.73--66.46) \\
         & VeinAttNet \cite{veinatnnet} & 98.10 (96.36--99.84) & 6.20 (2.53--9.87) & 20.70 (0.00--50.91) & 56.05 (21.04--91.05) & 97.46 (93.71--100.00) \\
         & Proposed Method & 99.53 (99.31--99.74) & 3.17 (2.12--4.22) & \textbf{53.52} (43.47--63.57) & 87.33 (80.87--93.78) & 99.86 (99.67--100.00) \\
        \hline
        $ACDE \rightarrow B$ & MCP \cite{mcp} & 92.90 & 12.31 & 0.04 & 3.29 & 77.65 \\
         & RLT \cite{rlt} & 90.31 & 13.76 & 0.00 & 0.04 & 57.20 \\
         & WLD \cite{wld} & 87.76 & 16.98 & 0.00 & 0.06 & 37.44 \\
         & ArcVein \cite{arcvein} & 91.74 (86.43--97.04) & 15.47 (9.11--21.83) & 1.49 (0.04--2.94) & 16.02 (5.75--26.28) & 72.07 (49.36--94.78) \\
         & LGFIN \cite{lgfin} & 92.55 (91.90--93.21) & 15.12 (14.27--15.97) & 5.31 (3.33--7.30) & 25.08 (22.97--27.20) & 75.64 (72.46--78.82) \\
         & FV-ViT \cite{vit} & 89.87 (87.55--92.20) & 17.74 (15.41--20.08) & 1.86 (0.77--2.95) & 13.17 (7.55--18.79) & 65.35 (55.79--74.92) \\
         & VeinAttNet \cite{veinatnnet} & \textbf{99.21} (98.92--99.50) & 4.07 (3.35--4.79) & \textbf{46.07} (32.67--59.48) & \textbf{82.77} (75.98--89.57) & 98.89 (98.49--99.28) \\
         & Proposed Method & 99.17 (98.88--99.45) & \textbf{3.91} (3.02--4.80) & 29.60 (27.53--31.66) & 76.96 (69.52--84.41) & \textbf{99.53} (99.12--99.93) \\
        \hline
        $ABDE \rightarrow C$ & MCP \cite{mcp} & 87.01 & 18.93 & 0.01 & 0.55 & 35.11 \\
         & RLT \cite{rlt} & 79.40 & 25.14 & 0.00 & 0.00 & 4.91 \\
         & WLD \cite{wld} & 83.07 & 22.87 & 0.00 & 0.08 & 20.58 \\
         & ArcVein \cite{arcvein} & 88.09 (84.76--91.42) & 19.74 (15.89--23.60) & 1.55 (0.11--2.98) & 8.65 (5.88--11.42) & 55.14 (42.22--68.07) \\
         & LGFIN \cite{lgfin} & 87.76 (86.73--88.80) & 18.51 (14.31--22.72) & 0.45 (0.00--1.27) & 4.20 (0.00--9.20) & 47.31 (14.97--79.66) \\
         & FV-ViT \cite{vit} & 88.76 (87.44--90.08) & 19.04 (17.80--20.28) & 1.16 (0.00--2.42) & 11.87 (6.76--16.97) & 62.83 (58.00--67.65) \\
         & VeinAttNet \cite{veinatnnet} & 86.93 (85.26--88.60) & 22.02 (19.66--24.37) & 6.07 (2.52--9.62) & 15.45 (10.92--19.98) & 50.11 (45.21--55.01) \\
         & Proposed Method & \textbf{97.78} (97.30--98.25) & \textbf{7.45} (6.52--8.38) & \textbf{16.82} (7.18--26.45) & \textbf{55.77} (50.67--60.86) & \textbf{95.39} (93.58--97.20) \\
        \hline
        $ABCE \rightarrow D$ & MCP \cite{mcp} & 95.77 & 9.90 & 3.83 & 21.88 & 90.35 \\
         & RLT \cite{rlt} & 88.17 & 16.51 & 0.00 & 0.07 & 34.81 \\
         & WLD \cite{wld} & 89.72 & 15.60 & 0.00 & 0.86 & 48.95 \\
         & ArcVein \cite{arcvein} & 84.16 (81.10--87.23) & 23.85 (20.43--27.27) & 0.56 (0.00--1.61) & 6.57 (3.28--9.86) & 49.40 (40.36--58.44) \\
         & LGFIN \cite{lgfin} & 93.65 (91.84--95.46) & 13.63 (11.45--15.82) & 13.83 (10.69--16.97) & 28.09 (19.74--36.44) & 78.94 (71.26--86.61) \\
         & FV-ViT \cite{vit} & 88.57 (86.33--90.80) & 19.26 (17.28--21.24) & 2.22 (0.00--4.91) & 12.93 (5.81--20.06) & 62.53 (55.11--69.96) \\
         & VeinAttNet \cite{veinatnnet} & 97.54 (96.99--98.10) & 8.17 (7.08--9.26) & 17.29 (0.00--38.14) & 57.93 (49.38--66.47) & 93.71 (91.71--95.72) \\
         & Proposed Method & \textbf{98.80} (98.43--99.17) & \textbf{5.20} (4.36--6.04) & \textbf{29.97} (15.63--44.31) & \textbf{72.23} (63.18--81.28) & \textbf{98.35} (97.68--99.02) \\
        \hline
        $ABCD \rightarrow E$ & MCP \cite{mcp} & 72.21 & 33.19 & \textbf{0.00} & 0.04 & 11.14 \\
         & RLT \cite{rlt} & 59.21 & 44.43 & \textbf{0.00} & 0.00 & 1.11 \\
         & WLD \cite{wld} & 69.67 & 35.45 & \textbf{0.00} & 0.02 & 8.51 \\
         & ArcVein \cite{arcvein} & 69.52 (66.76--72.29) & 35.56 (33.84--37.28) & \textbf{0.00} (0.00--0.00) & 1.11 (0.65--1.58) & 17.31 (15.56--19.07) \\
         & LGFIN \cite{lgfin} & 74.70 (73.03--76.37) & 33.16 (31.68--34.63) & \textbf{0.00} (0.00--0.00) & 5.69 (0.00--14.49) & 31.50 (23.59--39.41) \\
         & FV-ViT \cite{vit} & 73.16 (70.55--75.76) & 33.06 (29.81--36.32) & \textbf{0.00} (0.00--0.00) & 3.99 (0.02--7.96) & 31.22 (26.66--35.78) \\
         & VeinAttNet \cite{veinatnnet} & 88.43 (87.59--89.27) & 19.90 (18.18--21.62) & \textbf{0.00} (0.00--0.00) & \textbf{18.27} (7.66--28.88) & 61.01 (56.22--65.80) \\
         & Proposed Method & \textbf{89.98} (86.88--93.07) & \textbf{18.43} (15.70--21.16) & \textbf{0.00} (0.00--0.00) & 15.86 (2.85--28.86) & \textbf{64.68} (51.47--77.88) \\
        \hline
    \end{tabular}}
    \caption{Full-subject threshold-based operating-point results.}
    \label{tab:rebuttal_full_thresholds}
\end{table*}

\subsection{Dataset Quality Analysis and Performance Interpretation}
\label{sec:dataset_quality_analysis}

To better understand dataset-specific performance variations observed in Section~\ref{sec:results_discussion}, we analyse the intrinsic quality of the datasets using a set of handcrafted image-quality measures. The goal of this analysis is not to define a universal quality metric, but to provide consistent and interpretable indicators of structural clarity, continuity, and contrast across datasets.

For a grayscale image $I$ defined over pixel domain $\Omega$, we compute four complementary measures:
\begin{align}
Q_{\text{grad}}(I) &= \frac{1}{|\Omega|}\sum_{p \in \Omega} \sqrt{G_x(p)^2 + G_y(p)^2}, \\
Q_{\text{coh}}(I) &= \frac{1}{B}\sum_{b=1}^{B}
\left(
\frac{\sqrt{(j_{11}^{(b)}-j_{22}^{(b)})^2 + 4(j_{12}^{(b)})^2}}
{j_{11}^{(b)} + j_{22}^{(b)} + \varepsilon}
\right)^2, \\
Q_{\text{lap}}(I) &= \mathrm{Var}(\Delta I), \\
Q_{\text{con}}(I) &= \frac{1}{B}\sum_{b=1}^{B} \sigma_b,
\end{align}
where $G_x$ and $G_y$ denote Sobel gradients, $(j_{11}^{(b)}, j_{22}^{(b)}, j_{12}^{(b)})$ are structure-tensor components computed over block $b$, $\Delta I$ represents the Laplacian response, and $\sigma_b$ is the standard deviation within block $b$. In all cases, higher values indicate stronger structural visibility or contrast.

\begin{table}[ht!]
    \centering
    \resizebox{\columnwidth}{!}{
    \begin{tabular}{|l|c|c|c|c|c|}
        \hline
        Dataset & Gradient & Grad. Coherence & Laplacian & Contrast & Images/class \\
        \hline
        FV-300 (A) \cite{veinatnnet} & 0.0687 & 0.3675 & 0.0609 & 0.0311 & 89.57 \\
        MMCBNU (B) \cite{mmcbnu} & 0.0781 & 0.5398 & 0.0026 & 0.0392 & 10.00 \\
        FV-USM (C)\cite{fvusm} & 0.0312 & 0.2928 & 0.0009 & 0.0137 & 12.00 \\
        PolyU (D) \cite{polyu} & 0.0478 & 0.4605 & 0.0017 & 0.0249 & 5.84 \\
        VERA (E) \cite{vera} & 0.0476 & 0.1429 & 0.0074 & 0.0196 & 2.00 \\
        \hline
    \end{tabular}
    }
 \caption{Mean fingervein  image-quality indicators computed across datasets. The metrics quantify gradient strength, structural coherence, edge sharpness (Laplacian variance), and local contrast. Higher values indicate better visibility and continuity of vein structures. The number of images per subject is also reported to highlight dataset density, which impacts verification reliability.}
    \label{tab:rebuttal_quality}
\end{table}

The results in Table~\ref{tab:rebuttal_quality} provide useful insights into the observed performance trends. As observed in Tables~\ref{tab:rebuttal_half_thresholds} and~\ref{tab:rebuttal_full_thresholds}, handcrafted methods such as MCP and WLD outperform several deep learning baselines in the \(BCDE \rightarrow A\) setting. 
First, the $BCDE \rightarrow A$ setting corresponds to evaluation on FV-300, which exhibits relatively high structural quality and, importantly, a significantly larger number of samples per subject compared to other datasets. Under such controlled and high-quality conditions, handcrafted methods based on vessel enhancement and template matching remain highly competitive, as the vein structures are already well-defined and less affected by noise or variability. Therefore, the strong performance of handcrafted approaches in this setting does not contradict the overall findings, but rather highlights that simpler methods can perform well when the data distribution is favourable.

In contrast, the performance drop observed in the $ABCD \rightarrow E$ setting is consistent with both dataset characteristics and the quality analysis. VERA exhibits the lowest gradient-coherence among all datasets, indicating weaker structural continuity of vein patterns, and contains only two images per identity. This results in limited intra-class representation and a sparse score distribution during evaluation. Furthermore, the open-sensor acquisition setup introduces additional variability, increasing the domain shift relative to the training data.

As a result, the $ABCD \rightarrow E$ protocol represents a challenging low-quality, low-sample, and high-variability scenario. The reduced performance across all methods in this setting is therefore expected and reflects the inherent difficulty of the dataset rather than a limitation specific to the proposed approach.

Overall, this analysis clarifies that the observed performance variations are primarily driven by dataset characteristics, including image quality, sample density, and acquisition variability. These findings support a more precise interpretation of the experimental results, where the proposed method demonstrates strong generalisation across diverse conditions while remaining robust under challenging cross-dataset scenarios.

\textbf{Statistical significance:}
To validate that the observed improvements are not due to random variation, we performed paired statistical significance testing using a two-sided sign-flip permutation test on full-subject EER across matched runs. The proposed method shows statistically significant improvements over all competing deep learning baselines after Holm correction ($p < 10^{-3}$). Detailed results are provided in the supplementary material.

\section{Ablation Study}
\label{sec:ablation}

In this section, we present a consolidated ablation study to justify the main design choices of OpenVeinNet. The ablations analyse: (i) the kernel-size configuration of the DSConv stem, (ii) the number of Grapher Blocks in the backbone, (iii) the contribution of the proposed Centroid Angular Hybrid (CAH) loss compared with standard losses, (iv) its comparison against strong angular-margin losses, (v) the sensitivity of the balancing factor $\beta$, and (vi) the individual contributions of DSConv and graph-based modelling. Unless otherwise stated, the ablation experiments are conducted under the $ABDE \rightarrow C$ protocol (Full-subject), where FV-USM is used as the held-out evaluation dataset. Lower EER indicates better verification performance.

\subsection{Effect of DSConv Kernel Sizes}

\begin{table}[ht!]
    \centering
    \begin{tabular}{| c | c |}
        \hline 
        Kernel set & EER (\%)  \\
        \hline
        \hline
        (3, 3, 3) & 12.3351 \\
        (5, 5, 5) & 12.4453 \\
        (7, 7, 7) & 11.8358 \\
        (9, 9, 9) & 11.8376 \\
        (9, 7, 5) & 11.0605 \\
        (7, 5, 3) & 13.0697 \\
        (9, 5, 3) & 11.2155 \\
        \textbf{(9, 7, 3)} & \textbf{10.8235} \\
        \hline
    \end{tabular}
    \caption{Ablation on the kernel sizes used sequentially in the DSConv Blocks of the stem.}
    \label{tab:stemexps}
\end{table}

Table~\ref{tab:stemexps} shows the effect of different kernel-size configurations in the DSConv stem. Uniformly small kernels such as $(3,3,3)$ and $(5,5,5)$ are less effective, suggesting that very local receptive fields are insufficient to capture the extended curvilinear structure of finger veins. Uniformly large kernels improve the result, but do not provide the best performance. The best EER is obtained using the progressively decreasing configuration $(9,7,3)$, which achieves an EER of $10.82\%$.

This trend is technically meaningful. Larger kernels in the shallow layers help capture broader vascular structures and long vein segments, while smaller kernels in deeper layers refine local discriminative details. Therefore, the $(9,7,3)$ configuration provides a better balance between global vein continuity and fine local pattern representation.

\subsection{Effect of the Number of Grapher Blocks}

\begin{figure}[ht!]
\centering
\includegraphics[width=\linewidth]{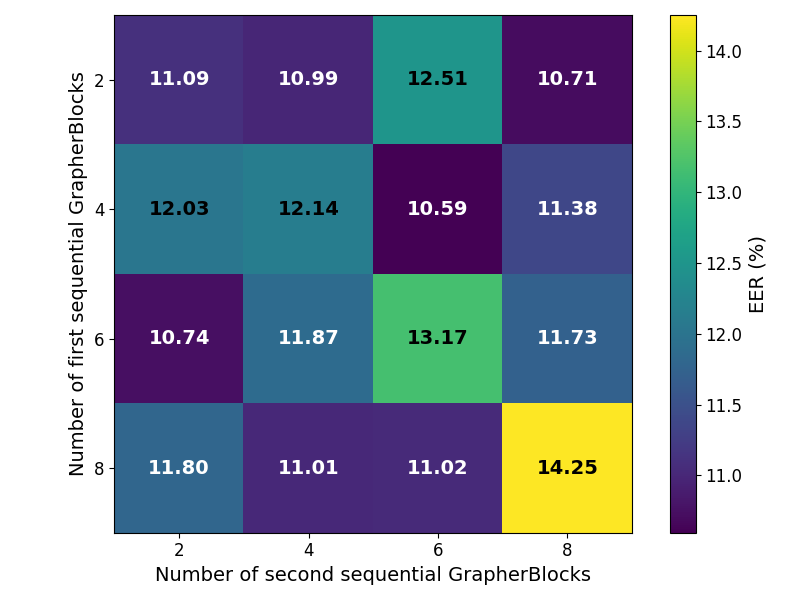}
\caption{Ablation on the number of Grapher Blocks in the backbone. The Y-axis indicates the number of Grapher Blocks in the first stage, while the X-axis indicates the number of Grapher Blocks in the second stage.}
\label{fig:bakcbone_graphers}
\end{figure}

The number of Grapher Blocks controls the depth of graph-based message passing in the backbone. A shallow graph backbone may not propagate sufficient relational information between vein regions, while an overly deep graph backbone may introduce redundancy, over-smoothing, or unnecessary computational cost. To identify an effective configuration, we evaluate different combinations of Grapher Blocks in the first and second backbone stages, selected from $\{2,4,6,8\}$.

As shown in Figure~\ref{fig:bakcbone_graphers}, the configuration with four Grapher Blocks in the first stage and six Grapher Blocks in the second stage achieves the best EER of $10.595\%$. This indicates that the model benefits from moderate graph depth: sufficient to capture long-range vein connectivity, but not so deep that feature representations become over-smoothed. This observation supports the final backbone design used in OpenVeinNet.

\subsection{Comparison with Standard Loss Functions}

\begin{table}[htbp]
\centering
\caption{Ablation study on standard loss functions under the $ABDE \rightarrow C$ protocol. Lower EER is better.}
\label{tab:loss_ablation}
\begin{tabular}{|l|c|}
\hline
\textbf{Loss Function} &  \textbf{EER (\%)} \\
\hline
MSE        &  26.68 \\
NLL        &  33.98 \\
\textbf{Proposed CAH Loss} & \textbf{7.44} \\
\hline
\end{tabular}
\end{table}

Table~\ref{tab:loss_ablation} compares the proposed CAH loss with standard MSE and NLL losses. The proposed loss achieves a substantially lower EER than both alternatives, confirming that regression-style or likelihood-based objectives are not sufficient for learning highly discriminative finger-vein embeddings. In contrast, the proposed loss explicitly promotes class separability and compact identity-specific embeddings, which are essential for verification.
\subsection{Sensitivity to Graph Neighbourhood Size $k$}
\begin{table}[t]
\centering
\caption{Sensitivity of the graph neighbourhood size $k$ in 
the two backbone stages on the \textit{ABDE}$\rightarrow$\textit{C} 
protocol. Lower EER is better.}
\begin{tabular}{ccc}
\hline
Stage I $k$ & Stage II $k$ & EER (\%) \\
\hline
9  & 9  & 10.37 \\
9  & 18 & 9.42  \\
18 & 18 & 10.48 \\
\textbf{18} & \textbf{9 (proposed)} & \textbf{7.44} \\
\hline
\end{tabular}
\label{tab:graph_k_sensitivity}
\end{table}
Table~\ref{tab:graph_k_sensitivity} reports the effect of the 
neighbourhood size $k$ used in the k-NN graph construction 
across the two backbone stages. Using the same value in both 
stages, whether $k{=}9$ or $k{=}18$, gives weaker performance 
than the proposed asymmetric configuration of $k{=}18$ in 
Stage~I and $k{=}9$ in Stage~II. The symmetric $k{=}18$ 
configuration performs the worst among all variants, suggesting 
that using a large neighbourhood in the later stage introduces 
noise from less relevant nodes as spatial resolution decreases. 
The proposed decreasing schedule allows Stage~I to aggregate 
broad vascular connectivity at higher resolution, while 
Stage~II refines more localised relational evidence at lower 
resolution. This mirrors the progressively decreasing kernel 
configuration used in the DSConv stem and reflects a consistent 
design principle: broader context in early stages, finer 
discrimination in later stages.
\subsection{Comparison with Strong Angular-Margin Losses}

\begin{table}[ht!]
    \centering
    \begin{tabular}{|l|c|}
        \hline
        Loss Function & Mean EER (\%) \\
        \hline
        ArcFace \cite{deng2019arcface} & 7.83 \\
        MagFace \cite{meng2021magface} & 8.20 \\
        AdaFace \cite{kim2022adaface} & 8.22 \\
        \textbf{Proposed CAH Loss} & \textbf{7.44} \\
        \hline
    \end{tabular}
    \caption{Loss ablation against strong angular-margin losses on the $ABDE \rightarrow C$ protocol. Lower EER is better.}
    \label{tab:rebuttal_loss_ablation}
\end{table}

To further validate the proposed loss, Table~\ref{tab:rebuttal_loss_ablation} compares CAH against strong angular-margin losses, including ArcFace, MagFace, and AdaFace. These losses are widely used for discriminative biometric representation learning. 
The proposed CAH loss achieves the lowest mean EER of 7.44\%, 
compared with 7.83\% for ArcFace, 8.20\% for MagFace, and 
8.22\% for AdaFace. The numerical differences are modest, 
reflecting the fact that all compared losses enforce angular 
separation to some degree. The distinction of the proposed 
loss lies in its explicit penalisation of intra-class scatter 
through the centroid alignment term, which complements the 
inter-class angular margin. In open-set verification, where 
decisions are made by thresholding cosine similarity against 
enrolled templates rather than through a fixed classifier, 
compact intra-class embeddings directly improve genuine score 
concentration and therefore TAR at strict operating thresholds. 
This is reflected in the TAR@FAR results in Tables~5 and~6, 
where the full model with CAH loss achieves competitive or 
higher TAR than the compared methods across most protocols.

\subsection{Sensitivity to the Balancing Factor $\beta$}

\begin{table}[ht!]
    \centering
    \begin{tabular}{|c|c|}
        \hline
        $\beta$ & EER (\%) \\
        \hline
        0.1 & 9.07 \\
        \textbf{0.3 (default)} & \textbf{7.44} \\
        0.5 & 10.60 \\
        0.9 & 11.57 \\
        \hline
    \end{tabular}
    \caption{Sensitivity of the proposed CAH loss to the balancing factor $\beta$ on the $ABDE \rightarrow C$ protocol. Lower EER is better.}
    \label{tab:rebuttal_beta_ablation}
\end{table}

Table~\ref{tab:rebuttal_beta_ablation} reports the sensitivity of the proposed loss to the angular balancing factor $\beta$. The best performance is achieved at $\beta=0.3$, with an EER of $7.44\%$. When $\beta$ is too small, such as $0.1$, the angular component is under-weighted and the model does not fully benefit from angular supervision. Conversely, when $\beta$ is too large, such as $0.5$ or $0.9$, the angular term dominates the optimisation and degrades performance.

This confirms that the angular term is useful, but must be balanced carefully. The default value $\beta=0.3$ provides the best trade-off between classification supervision and angular regularisation, supporting the choice used in the final model.

\subsection{Component-Level Ablation of DSConv and Graph Modelling}

\begin{table}[ht!]
    \centering
    \begin{tabular}{|l|c|}
        \hline
        Variant & Mean EER (\%) \\
        \hline
        Standard stem + no graph & 13.59 \\
        DSConv stem + no graph & 10.96 \\
        Standard stem + graph backbone & 11.64 \\
        \textbf{Full model (DSConv + graph backbone)} & \textbf{7.44} \\
        \hline
    \end{tabular}
    \caption{Component ablation isolating the contributions of DSConv and graph modelling on the $ABDE \rightarrow C$ protocol. Lower EER is better.}
    \label{tab:rebuttal_component_ablation}
\end{table}

Table~\ref{tab:rebuttal_component_ablation} isolates the contribution of the DSConv stem and the graph-based backbone. The standard convolutional stem without graph modelling gives the weakest performance, with an EER of $13.59\%$. Replacing the standard stem with DSConv reduces the EER to $10.96\%$, showing that adaptive tubular feature extraction is beneficial for finger-vein representation. Similarly, adding the graph backbone to a standard stem improves the EER to $11.64\%$, demonstrating that graph-based relational modelling also contributes positively.

The full model achieves an EER of 7.44\%, compared with 
13.59\% for the standard baseline, 10.96\% for DSConv alone, 
and 11.64\% for the graph backbone alone. The combined 
improvement is larger than the sum of the individual gains 
from each component.  This suggests that 
the two components interact synergistically: graph-based 
relational reasoning is most effective when node features 
are already geometrically aligned with vein structures, a 
condition that DSConv satisfies by design but standard 
convolution does not. The combination therefore produces 
gains that neither component achieves in isolation.

\subsection{Overall Ablation Analysis}

The ablation results provide strong evidence for the design of OpenVeinNet. The stem analysis shows that a progressively decreasing DSConv kernel configuration, $(9,7,3)$, is most effective for capturing both broad and fine vein structures. The Grapher Block analysis confirms that a moderate graph depth, using four Grapher Blocks in the first stage and six in the second, provides the best balance between relational modelling and representation stability.

The loss ablations further show that the proposed CAH loss is a key contributor to the final performance. It substantially outperforms standard losses and also achieves the lowest EER among the evaluated angular-margin alternatives, including ArcFace, MagFace, and AdaFace. The $\beta$ sensitivity study confirms that the angular component is beneficial when properly balanced, with $\beta=0.3$ giving the best result.

Finally, the component-level ablation directly addresses whether the architecture is merely a combination of existing modules. The results show that both DSConv and graph modelling independently improve performance, and their combination provides the strongest result. This demonstrates that OpenVeinNet benefits from a synergistic interaction between local tubular feature extraction, global graph-based relational modelling, and angularly constrained embedding learning.

Overall, these ablation studies confirm that each major component of the proposed framework contributes meaningfully to the final verification performance, thereby validating the architectural and loss-design choices of OpenVeinNet.

\section{Computational Cost and Inference Latency}
\label{sec:efficiency}

In addition to verification performance, practical deployment of biometric systems requires careful consideration of computational cost and inference latency. While the proposed \textit{OpenVeinNet} is designed primarily for robust open-set verification, it is equally important to quantify its efficiency characteristics and understand the associated trade-offs relative to existing methods.

To this end, we benchmark all learned approaches under a unified experimental protocol. Specifically, we evaluate each method using its held-out FV-300 checkpoint, a batch size of $1$, and repeated forward passes on both CPU and GPU platforms. We report the number of learnable parameters as a proxy for model size, FLOPs as an approximate measure of computational complexity, and mean single-image latency and throughput. Handcrafted methods (MCP, RLT, WLD) are excluded due to the absence of a unified neural inference pipeline. For VeinAttNet, FLOPs are reported as \texttt{n/a}, as its available implementation is in MATLAB and not directly compatible with the same profiling tools.

\begin{table*}[ht!]
    \centering
    \scriptsize
    \resizebox{\textwidth}{!}{
    \begin{tabular}{|l|c|c|c|c|c|c|}
        \hline
        Method & Params (M) & FLOPs (G) & CPU Lat. (ms) & CPU Thr. (img/s) & GPU Lat. (ms) & GPU Thr. (img/s) \\
        \hline
        ArcVein \cite{arcvein} & 7.83 & 2.45 & 7.09 & 141.10 & \textbf{0.94} & \textbf{1067.42} \\
        LGFIN \cite{lgfin} & 6.79 & 7.51 & 45.48 & 21.99 & 3.84 & 260.42 \\
        FV-ViT \cite{vit} & \textbf{0.36} & \textbf{0.03} & \textbf{5.02} & \textbf{199.37} & 1.74 & 574.79 \\
        VeinAttNet \cite{veinatnnet} & \textbf{0.09} & n/a & 11.24 & 88.96 & 10.63 & 94.08 \\
        Proposed Method & 4.76 & 5.07 & 55.82 & 17.92 & 13.00 & 76.94 \\
        \hline
    \end{tabular}}
    \caption{Runtime and computational-cost comparison on FV-300 using batch size $1$. Lower latency is better; higher throughput is better.}
    \label{tab:rebuttal_runtime}
\end{table*}

Table~\ref{tab:rebuttal_runtime} summarises the computational characteristics of the evaluated methods. Several key observations can be drawn.

\textbf{Efficiency-oriented models.}
Lightweight architectures such as FV-ViT exhibit the lowest computational footprint, with only $0.36$M parameters and $0.03$ GFLOPs, resulting in the fastest CPU latency ($5.02$ ms) and highest CPU throughput. Similarly, ArcVein achieves the fastest GPU latency ($0.94$ ms) and highest throughput, making it well suited for high-throughput scenarios. However, as demonstrated in Section~\ref{sec:results_discussion}, these efficiency gains come at the cost of reduced verification robustness, particularly under cross-dataset and open-set conditions.

\textbf{Proposed method: accuracy-driven design.}
The proposed \textit{OpenVeinNet} adopts a more expressive architecture, resulting in a moderate computational footprint of $4.76$M parameters and $5.07$ GFLOPs. This leads to higher latency (55.82 ms on CPU and 13.00 ms on GPU) compared to lightweight baselines. Nevertheless, the model remains computationally practical, especially in GPU-enabled or server-side deployment scenarios. Importantly, OpenVeinNet is still more compact than several CNN-based baselines (e.g., ArcVein and LGFIN) in terms of parameter count, while delivering substantially stronger verification performance.

\textbf{Justification of computational cost.}
The increased computational cost is a direct consequence of the architectural components that enable improved verification performance. The DSConv layers introduce adaptive sampling mechanisms that align receptive fields with curvilinear vein structures, improving feature quality at the expense of additional computation. The Grapher Blocks further incorporate dynamic graph construction and message passing, allowing the model to capture long-range topological dependencies that cannot be effectively modelled by standard convolutions. While these operations are computationally more demanding, they contribute directly to the discriminative strength and generalisability of the learned embeddings.

In addition, the proposed Centroid Angular Hybrid Loss improves embedding separability in angular space, which is particularly beneficial for open-set verification. As shown in Section~\ref{sec:results_discussion}, this results in consistently lower EER and higher TAR, especially under strict low-FAR operating conditions that are critical for real-world biometric systems.

\textbf{Accuracy-efficiency trade-off.}
Overall, the results highlight a clear and expected accuracy--efficiency trade-off. Methods such as FV-ViT and ArcVein achieve lower latency but exhibit weaker generalisation and reduced robustness under domain shift. In contrast, OpenVeinNet prioritises discriminative representation learning and open-set reliability, leading to improved verification performance at the cost of moderate computational overhead.

From a deployment perspective, this trade-off is well justified. In security-critical applications, such as financial authentication and identity verification, robustness and low error rates are typically prioritised over minimal latency. At the same time, the proposed model remains sufficiently efficient for practical deployment, particularly when GPU acceleration is available.

In summary, OpenVeinNet achieves a favourable balance between computational cost and verification performance, offering strong cross-dataset and open-set robustness while maintaining practical inference efficiency for GPU-enabled deployment.
\begin{figure*}[ht!]
\centering
\includegraphics[width=0.9\textwidth]{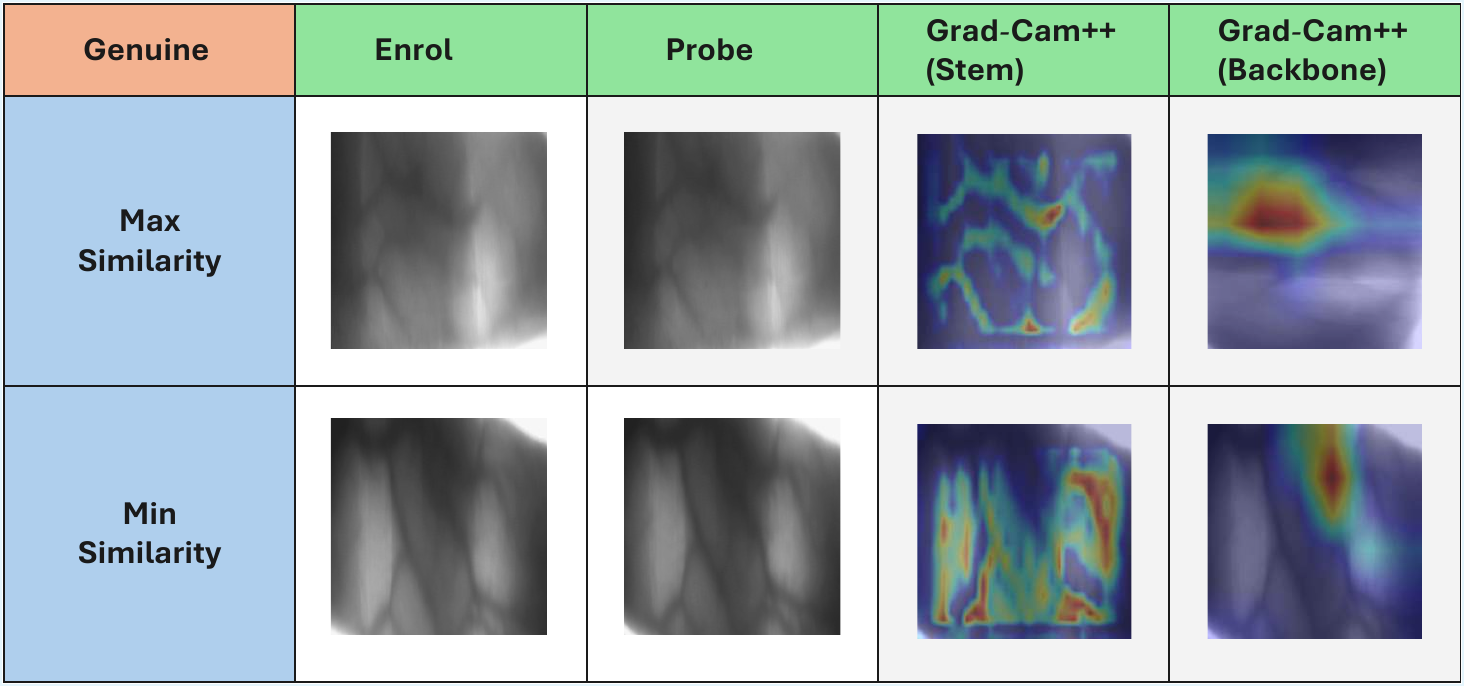}
\caption{Grad-CAM++~\cite{gradcamplusplus}-based interpretation of the proposed model for finger vein images from the same identity, representing a genuine comparison.}
\label{fig:explainaibility_genuine}
\end{figure*}

\begin{figure*}[ht!]
\centering
\includegraphics[width=0.9\textwidth]{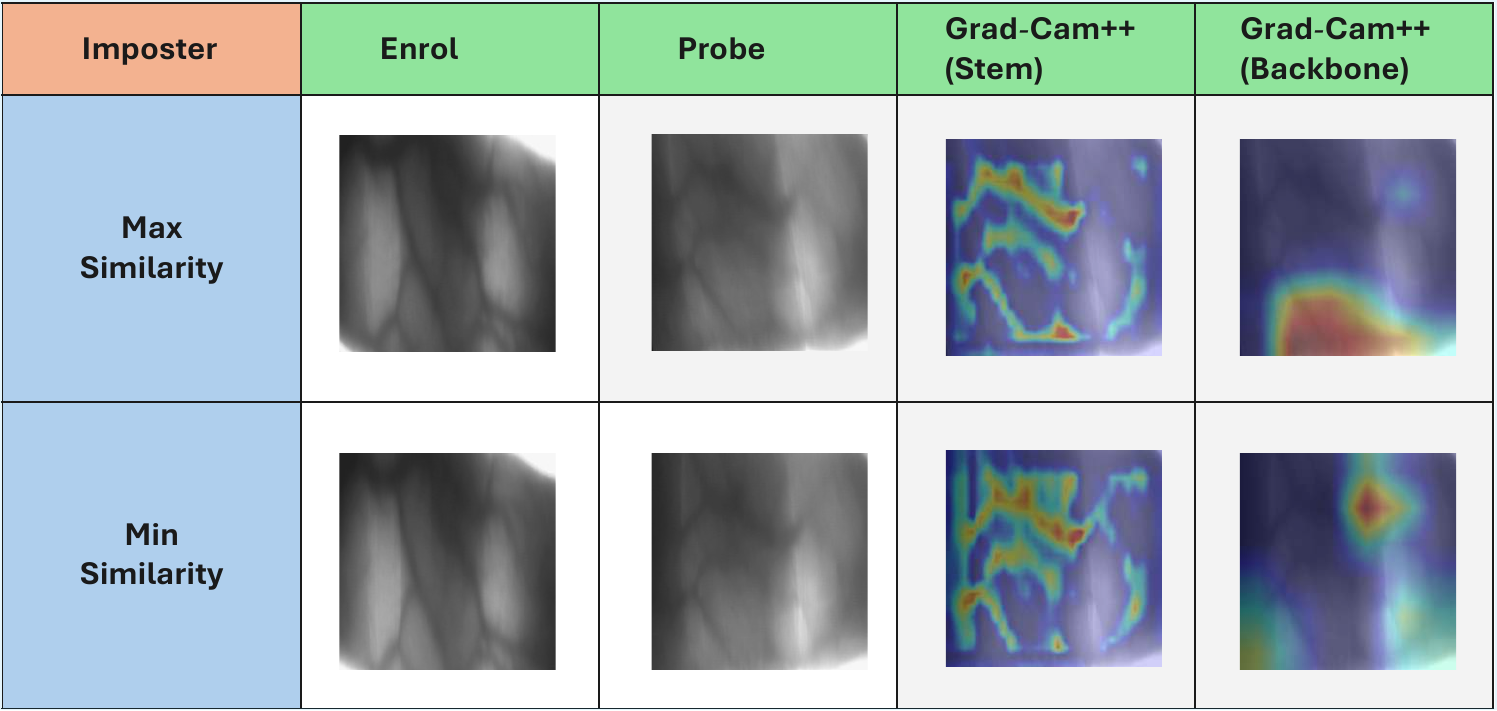}
\caption{Grad-CAM++~\cite{gradcamplusplus}-based interpretation of the proposed model for finger vein images from different identities, representing an impostor comparison.}
\label{fig:explainaibility_imposter}
\end{figure*}

\section{Interpretation of the Proposed Method}
\label{sec:interpretation}

Interpretability is important for understanding the decision behaviour of deep learning-based biometric systems, particularly in security-sensitive verification tasks. In this work, we employ Grad-CAM++~\cite{gradcamplusplus} to visualise the image regions that contribute most strongly to the proposed model's embedding responses. This analysis helps verify whether \textit{OpenVeinNet} focuses on meaningful vascular structures rather than irrelevant background or sensor-specific artefacts.

We analyse two verification cases: (i) \textbf{genuine comparisons}, where the enrolment and probe images belong to the same identity, and (ii) \textbf{impostor comparisons}, where the enrolment and probe images belong to different identities. For each case, we examine both high-similarity and low-similarity pairs. Grad-CAM++ activation maps are computed separately for the stem and backbone components to understand how local tubular feature extraction and graph-based feature aggregation contribute to the final representation.

Figures~\ref{fig:explainaibility_genuine} and~\ref{fig:explainaibility_imposter} show that the stem activations are strongly aligned with visible vein structures in the input images. This behaviour is expected because the DSConv stem is designed to follow curvilinear and tubular patterns. In genuine high-similarity cases, the activations are more concentrated around consistent vascular regions, indicating that the model relies on stable vein evidence for matching. In low-similarity or impostor cases, the responses become more diffuse or shift towards less consistent regions, reflecting weaker correspondence between the compared samples.

The backbone activations show a different behaviour. Rather than focusing only on local vein texture, the backbone highlights broader regions that appear to encode higher-level identity-specific information. This is consistent with the role of the Grapher Blocks, which aggregate relational information between neighbouring and non-local vein regions. Therefore, the interpretation results support the intended architectural design: the stem captures fine-grained tubular vein morphology, while the graph-based backbone refines these features into a more discriminative identity representation.

Overall, the Grad-CAM++ analysis provides qualitative evidence that \textit{OpenVeinNet} bases its verification decisions on meaningful vascular patterns and their higher-level structural relationships. This supports the reliability of the proposed architecture and helps explain its strong performance under cross-dataset and open-set verification settings.
\section{Conclusion}
\label{sec:conclusion}
Finger vein biometrics offer inherent resistance to spoofing and therefore represent a highly reliable modality for secure access control applications. In this work, we introduced \textbf{OpenVeinNet}, a novel framework for open-set finger vein verification that explicitly models both local vascular morphology and global structural relationships.
The proposed architecture integrates a Dynamic Snake Convolution (DSConv)-based \emph{Stem} for adaptive extraction of curvilinear vein patterns, with a graph-based \emph{Backbone} that captures long-range dependencies through relational feature aggregation. In addition, the proposed \emph{Centroid Angular Hybrid (CAH) Loss} enhances the discriminative capability of the embedding space by jointly enforcing intra-class compactness and inter-class angular separability, which is particularly beneficial in open-set scenarios.
Extensive cross-dataset experiments conducted under both unknown-rejection and full-subject verification protocols demonstrate that \textbf{OpenVeinNet} consistently demonstrates strong generalisation  compared with state-of-the-art methods across multiple benchmarks. The results show significant improvements in EER and TAR, especially under strict low-FAR operating conditions, confirming the robustness of the learned representations in challenging real-world scenarios.
Furthermore, the computational analysis highlights a favourable accuracy--efficiency trade-off, where the moderate increase in computational cost is justified by substantial gains in verification performance and generalisation. The interpretability analysis using Grad-CAM++ further validates that the model focuses on meaningful vascular structures, reinforcing the reliability of its decision-making process.
Overall, the proposed framework provides a principled and effective solution for open-set finger vein verification. Future work will explore lightweight variants of the architecture for edge deployment and investigate its applicability to presentation attack detection and multimodal biometric systems.


\bibliographystyle{IEEEtran}
\bibliography{bibliography}
\vskip -3\baselineskip plus -1fil
\begin{IEEEbiography}[{\includegraphics[width=1in,height=1.25in,clip,keepaspectratio]{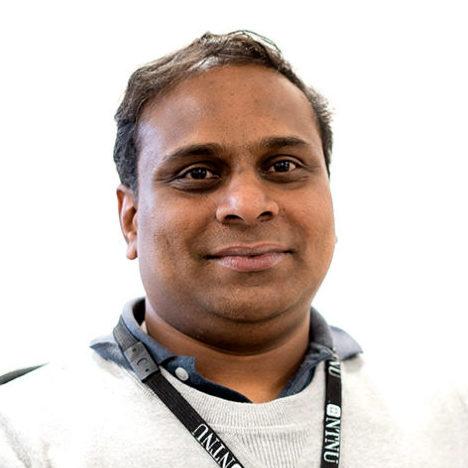}}]{Raghavendra Ramachandra}
received his Ph.D. in computer science and technology from the University of Mysore, India, and Télécom SudParis, France, in 2010. He is currently a Full Professor at the Department of Information Security and Communication Technology, Norwegian University of Science and Technology (NTNU), Gjøvik, Norway, and Scientific Director of the SAFE Centre. His research interests include biometrics, presentation attack detection, morphing attack detection, multimodal fusion, deep learning, and image/video analysis. He has contributed to several EU, IARPA, and national research projects and holds multiple patents in biometric security. He is actively involved in ISO/IEC JTC1 SC37 biometric standardization and serves as an editor of ISO/IEC 24722 on multimodal biometrics. He is a Senior Member of the IEEE, has received several best paper awards, and currently serves as Vice President for Finances of the IEEE Biometrics Council.
\end{IEEEbiography}
\begin{IEEEbiography}[{\includegraphics[width=1in,height=1.25in,clip,keepaspectratio]{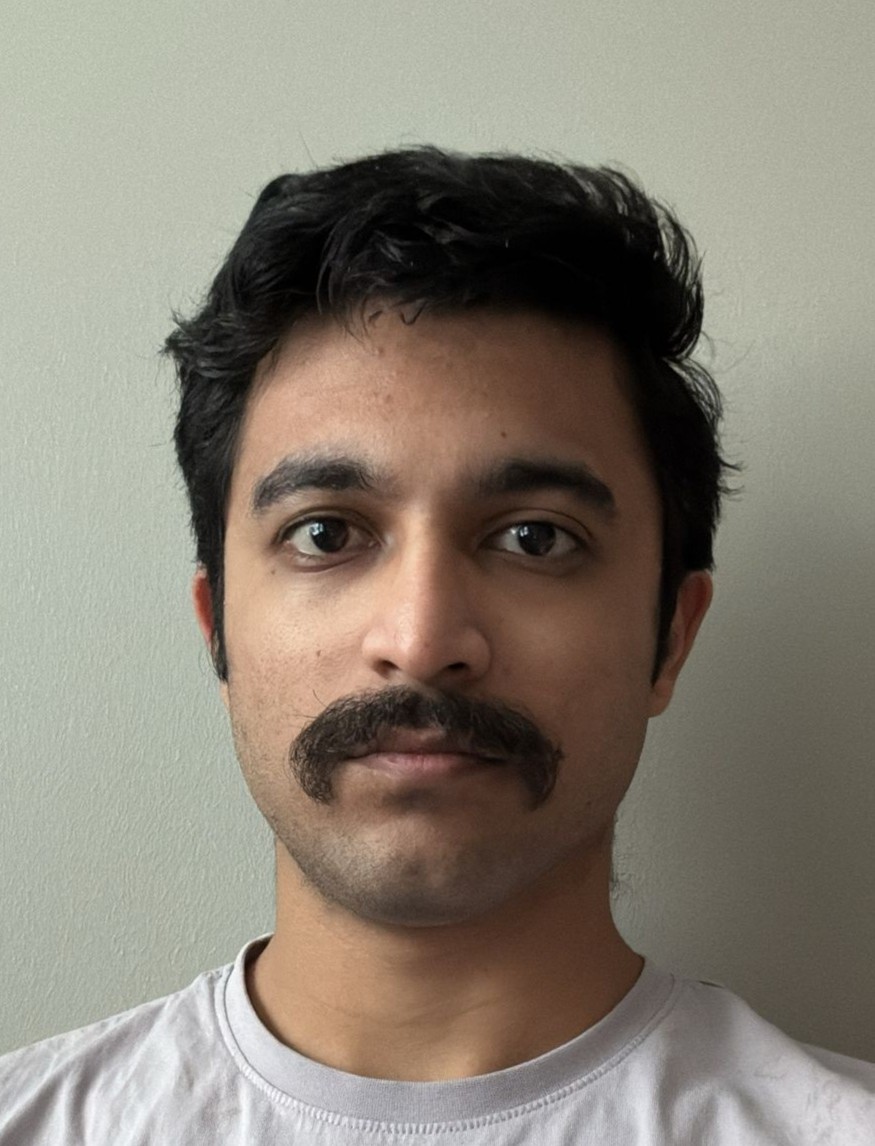}}]{Sushrut Patwardhan}
is a PhD student at the Norwegian University of Science and Technology (NTNU) and a data scientist at Mobai AS. His research focuses on deep learning applications in face biometrics, where he explores innovative solutions for identity verification and security. He is keen to investigate diverse deep learning architectures, constantly seeking to improve the accuracy, efficiency, and robustness of biometric systems.
\end{IEEEbiography}

\clearpage

\begin{center}
    {\LARGE \bf Supplementary Results \\ OpenVeinNet: Robust Open-Set Finger Vein Verification with Dynamic Snake Convolution and Graph Learning \par}
\end{center}

This supplementary material provides additional experimental evidence and
implementation details for OpenVeinNet, a finger-vein verification framework
based on dynamic snake convolution and graph-based representation learning.
We first report paired statistical significance tests on full-subject
verification results using seed-matched leave-one-dataset-out runs. Exact
two-sided sign-flip permutation tests with Holm correction show that the
equal error rate reductions obtained by OpenVeinNet are statistically
significant against ArcVein, LGFIN, FV-ViT, and VeinAttNet. We then present
intra-database open-set verification experiments on FV-300, FV-USM, and
MMCBNU using subject-disjoint identity partitions. OpenVeinNet achieves the
lowest equal error rate on all three datasets, with values of $0.39\%$,
$7.06\%$, and $4.07\%$, respectively, while maintaining strong true accept
rates across practical false accept rate operating points. To support
reproducibility, we document the training configurations, optimisation
settings, input resolutions, losses, and augmentation strategies used for all
learned baselines, together with the identity-level train--validation and
held-out splits. Finally, t-SNE visualisations and detection error trade-off
curves are used to examine the learned embedding space and verification
behaviour. The results show compact and separable embeddings for higher
quality datasets, increased overlap for low-sample and high-variability
datasets, and favourable false match and false non-match trade-offs across
evaluation protocols. Overall, the supplementary results support the
statistical reliability, reproducibility, and open-set verification behaviour
of OpenVeinNet.


\section{Statistical Significance Analysis}
\label{sec:statistical_significance}

To complement the mean and 95\% confidence interval reporting in the main paper, we perform formal pairwise statistical significance testing between the proposed method and competing approaches.

We conduct paired significance testing on the \emph{full-subject verification EER} results using two-sided exact sign-flip permutation tests over matched leave-one-dataset-out runs. For each comparison, pairing is performed across seed-matched runs obtained from the five evaluation protocols. Since lower EER indicates better performance, a positive mean $\Delta$EER corresponds to an average reduction in EER in favour of the proposed method.
To account for multiple comparisons, Holm correction is applied to the resulting $p$-values.

\begin{table*}[ht!]
    \centering
    \begin{tabular}{|l|c|c|c|c|c|}
        \hline
        Comparison & $n$ & Mean $\Delta$EER (pp) & Holm-adjusted $p$ & Wilcoxon $p$ & Significant \\
        \hline
        Proposed vs ArcVein & 20 & 15.17 & $7.63 \times 10^{-6}$ & $1.91 \times 10^{-6}$ & Yes \\
        Proposed vs LGFIN & 20 & 13.05 & $7.63 \times 10^{-6}$ & $1.91 \times 10^{-6}$ & Yes \\
        Proposed vs FV-ViT & 20 & 13.66 & $7.63 \times 10^{-6}$ & $1.91 \times 10^{-6}$ & Yes \\
        Proposed vs VeinAttNet & 20 & 4.44 & $9.54 \times 10^{-5}$ & $1.68 \times 10^{-4}$ & Yes \\
        \hline
    \end{tabular}
    \caption{Paired statistical significance testing on full-subject EER. A positive mean $\Delta$EER indicates that the proposed method achieves lower EER than the comparator. Holm-adjusted $p$ values are computed from the exact paired sign-flip permutation test, while the Wilcoxon column reports the paired two-sided signed-rank test.}
    \label{tab:statistical_significance}
\end{table*}

The results show (see Table \ref{tab:statistical_significance}) that all pairwise comparisons are statistically significant after Holm correction. This confirms that the observed improvements of OpenVeinNet over competing methods are not only reflected in mean performance, but are also consistent across matched evaluation runs.

\section{Intra-Database Open-Set Evaluation}
\label{sec:intra_database_open_set}
\subsection{Intra-Dataset Open-Set Evaluation}

To complement the cross-dataset evaluation, we further conduct intra-database open-set experiments on FV-300, FV-USM, and MMCBNU. These datasets provide a sufficient number of samples per identity to support training and evaluation on identity-disjoint subsets within the same database. For each dataset and statistical seed, the first 80\% of the sorted identities are used for model development, while the remaining 20\% of identities are held out entirely and used only for open-set verification. During evaluation, embeddings are extracted for the held-out identities, and genuine and impostor scores are computed from within-identity and cross-identity image pairs among these unseen subjects. PolyU and VERA are not included in this experiment because they contain only four and two samples per identity, respectively, which is insufficient for a reliable intra-database learning and held-out evaluation split under the proposed protocol.

\begin{table*}[ht!]
\centering
\scriptsize
\setlength{\tabcolsep}{3.5pt}
\renewcommand{\arraystretch}{1.08}
\caption{Intra-database open-set verification results with 95\% confidence intervals over five statistical seeds. The experiments are reported on FV-300, FV-USM, and MMCBNU, where sufficient samples per identity are available for reliable intra-database training and evaluation.}
\label{tab:intra_open_set_results}
\resizebox{\textwidth}{!}{%
\begin{tabular}{llccccc}
\hline
Dataset & Algorithm & AUC (\%) & EER (\%) & TAR@FAR$=0.1\%$ (\%) & TAR@FAR$=1\%$ (\%) & TAR@FAR$=10\%$ (\%) \\
\hline
\multirow{6}{*}{FV-300}
  & MCP  & 99.96 & 0.42 & 95.32 & \textbf{100.00} & \textbf{100.00} \\
  & RLT  & 99.92 & 0.74 & 78.14 & 99.70 & \textbf{100.00} \\
& ArcVein & 96.36 (94.01--98.72) & 11.03 (5.95--16.11) & 54.26 (45.30--63.22) & 67.25 (57.34--77.15) & 89.13 (81.94--96.31) \\
& LGFIN & 99.99 (99.98--100.00) & 0.43 (0.09--0.78) & 98.42 (96.98--99.86) & 99.77 (99.44--100.00) & 100.00 (100.00--100.00) \\
& FV-ViT & 96.55 (95.04--98.06) & 9.60 (7.59--11.61) & 29.05 (7.24--50.86) & 48.83 (28.26--69.40) & 90.48 (85.25--95.70) \\
& VeinAttNet & 99.73 (99.63--99.83) & 2.43 (1.81--3.05) & 63.32 (47.63--79.01) & 93.71 (90.81--96.60) & 99.90 (99.86--99.94) \\
& Proposed Method & \textbf{99.99} (99.98--100.00) & \textbf{0.39} (0.06--0.72) & \textbf{98.66} (96.85--100.00) & 99.77 (99.41--100.00) & \textbf{100.00 (100.00--100.00)}\\
\hline
\multirow{6}{*}{FV-USM}
& MCP  & 86.64 & 18.98 & 0.01 & 0.62 & 30.20 \\
& RLT  & 79.69 & 24.76 & 0.00 & 0.01 & 5.35 \\
& ArcVein & 65.05 (61.48--68.62) & 38.01 (35.77--40.24) & 0.69 (0.00--1.64) & 2.69 (1.08--4.30) & 18.54 (15.01--22.07) \\
& LGFIN & 95.60 (94.71--96.48) & 11.25 (9.89--12.61) & 28.91 (7.57--50.25) & 45.42 (27.85--62.99) & 87.17 (83.96--90.39) \\
& FV-ViT & 82.90 (79.16--86.65) & 26.25 (23.52--28.98) & 5.72 (0.00--14.46) & 15.93 (3.59--28.26) & 45.14 (33.70--56.58) \\
& VeinAttNet & 83.85 (77.20--90.51) & 25.60 (20.08--31.11) & 10.73 (3.40--18.05) & 19.45 (6.84--32.06) & 50.97 (34.37--67.56) \\
& Proposed Method & \textbf{98.26 (97.48--99.04)} & \textbf{7.06 (5.40--8.72)} & \textbf{53.38 (41.30--65.46)} & \textbf{73.66 (67.37--79.94)} & \textbf{95.33 (92.58--98.08)} \\
\hline
\multirow{6}{*}{MMCBNU}
& MCP  & 93.04 & 12.24 & 0.04 & 3.84 & 77.91 \\
& RLT  & 90.44 & 13.76 & 0.00 & 0.08 & 59.24 \\
& ArcVein & 78.51 (72.71--84.30) & 29.70 (24.39--35.00) & 4.79 (0.83--8.75) & 9.20 (3.09--15.30) & 30.04 (20.11--39.96) \\
& LGFIN & 99.19 (98.96--99.42) & 4.33 (3.48--5.17) & \textbf{52.17 (41.93--62.41)} & 84.43 (80.43--88.44) & 98.50 (97.78--99.22) \\
& FV-ViT & 92.20 (91.14--93.27) & 16.25 (14.99--17.51) & 21.14 (14.92--27.36) & 40.06 (35.14--44.97) & 74.98 (71.56--78.40) \\
& VeinAttNet & 92.52 (87.99--97.04) & 15.00 (9.46--20.54) & 10.71 (0.21--21.20) & 39.39 (21.50--57.28) & 79.25 (65.82--92.68) \\
& Proposed Method & \textbf{99.31 (99.13--99.50)} & \textbf{4.07 (3.43--4.71)} & 49.24 (30.59--67.90) & \textbf{85.92 (82.15--89.69)} & \textbf{98.91 (98.45--99.37)} \\
\hline
\end{tabular}}
\end{table*}

Table~\ref{tab:intra_open_set_results} reports the intra-database open-set verification results. On FV-300, all strong methods perform well due to the high image quality and large number of samples per identity. Nevertheless, the proposed method achieves the lowest EER of $0.39\%$ and the highest TAR at FAR$=0.1\%$, while matching the best TAR at FAR$=1\%$ and FAR$=10\%$. This confirms that OpenVeinNet remains highly competitive even when the intra-database setting is favourable to several baseline methods.

On FV-USM, the advantage of the proposed method becomes more evident. OpenVeinNet achieves an AUC of $98.26\%$ and an EER of $7.06\%$, substantially improving over the second-best LGFIN result of $11.25\%$ EER. The proposed method also provides the highest TAR across all operating points, including $53.38\%$ at FAR$=0.1\%$ and $73.66\%$ at FAR$=1\%$. These results indicate that the proposed representation is more robust when intra-class variability and acquisition differences are stronger.

On MMCBNU, the proposed method again achieves the best AUC and EER, with $99.31\%$ AUC and $4.07\%$ EER. LGFIN obtains a slightly higher TAR at the strictest FAR$=0.1\%$, but OpenVeinNet achieves the best TAR at FAR$=1\%$ and FAR$=10\%$, showing a more favourable operating behaviour across practical thresholds. This suggests that the proposed combination of DSConv, graph-based modelling, and CAH loss produces embeddings that remain discriminative across both strict and relaxed operating points.

Overall, the intra-database results support the conclusions from the cross-dataset experiments. The proposed method performs strongly not only when evaluated across different datasets, but also when trained and tested within the same database under an open-set protocol. This additional experiment directly confirms that OpenVeinNet is effective for conventional intra-database open-set verification, while the cross-dataset experiments demonstrate its generalisation ability under stronger domain shift.
\section{Baseline Training Details and Experimental Fairness}
\label{sec:supp_baseline_training}

\subsection{Overview}

To ensure transparency and reproducibility, we provide detailed training configurations for all learned baselines used in the experiments. Unlike many prior works that report results directly from the literature, all deep learning baselines in this work were \textbf{retrained by us} under a unified evaluation protocol.

Specifically, ArcVein, LGFIN, FV-ViT, and the proposed method were implemented and trained in a common PyTorch framework, while VeinAttNet was independently retrained in MATLAB using the published architecture within our codebase. Handcrafted methods (MCP, RLT, and WLD) do not involve learning and were executed as fixed pipelines.

\section{Shared Experimental Protocol}

All learned methods were trained and evaluated under the same \textbf{leave-one-dataset-out} protocol. For each experiment:
\begin{itemize}
    \item Four datasets were used for training, and one dataset was held out for testing.
    \item The same statistical seeds were used across all methods to ensure matched runs.
    \item A common evaluation pipeline was used for computing AUC, EER, and TAR@FAR.
\end{itemize}

For PyTorch-based models, input images were converted from grayscale to three channels and processed using a shared data pipeline with the following augmentations:
\begin{itemize}
    \item Random horizontal and vertical flips
    \item Random autocontrast ($p = 0.05$)
    \item Random rotations within $\pm 45^\circ$
    \item Resizing to method-specific input resolution
\end{itemize}

VeinAttNet followed its original MATLAB-based augmentation strategy, including random reflections, translations (up to $\pm 30$ pixels), rotations, and scaling.

Importantly, \textbf{no dataset-specific tuning was performed on the held-out test set}. Each method used a fixed training configuration across all splits, except for the unavoidable change in the number of source classes.

\subsection{Baseline Training Configurations}

Table~\ref{tab:rebuttal_baseline_training} summarises the training configurations for all learned methods.

\begin{table*}[ht!]
    \centering
    \scriptsize
    \resizebox{\textwidth}{!}{
    \begin{tabular}{|l|l|c|c|c|c|c|l|}
        \hline
        Method & Training setup & Optimizer & LR & Epochs & Batch & Input & Loss \\
        \hline
        MCP / RLT / WLD & Handcrafted pipelines & n/a & n/a & n/a & n/a & Native & No learning \\
        ArcVein & Retrained (PyTorch) & SGD (Nesterov 0.9) & $1\times10^{-2}$ ($\div 10$ @30) & 100 & 64 & $224\times224$ & Arc-cosine + softmax ($\lambda=0.01$) \\
        LGFIN & Retrained (PyTorch) & AdamW & $1\times10^{-4}$ & 50 & 8 & $147\times147$ & Cross-entropy \\
        FV-ViT & Retrained (PyTorch) & AdamW & $1\times10^{-4}$ & 50 & 64 & $224\times224$ & Cross-entropy \\
        VeinAttNet & Retrained (MATLAB) & Adam & $1\times10^{-3}$ & 50 & 128 & $224\times224$ & Softmax classification \\
        Proposed & Retrained (PyTorch) & AdamW & $1\times10^{-3}$ & 150 & 128 & $224\times224$ & CAH ($\beta=0.3$) \\
        \hline
    \end{tabular}}
    \caption{Training configurations for the proposed method and all learned baselines.}
    \label{tab:rebuttal_baseline_training}
\end{table*}

\subsection{Fairness of Comparison}

The comparison is fair in the sense that all learned methods were:
\begin{itemize}
    \item Retrained under the same cross-dataset protocol,
    \item Evaluated using identical metrics and code,
    \item Compared across matched statistical runs.
\end{itemize}

At the same time, the comparison does not enforce identical hyperparameters across all methods. Each baseline retains its architecture-specific design choices, including input resolution, optimizer type, learning rate schedule, and loss formulation. For example, ArcVein is trained using its original SGD/Nesterov schedule rather than being forced into an Adam-based setup.

Enforcing identical hyperparameters across fundamentally different architectures would introduce a different form of bias. Therefore, we adopt a \textbf{method-consistent but protocol-aligned} evaluation strategy, which preserves the intended behaviour of each baseline while ensuring comparability across methods.

\section{Subject-Disjoint Train--Validation Split}
\label{sec:supp_trainval_split}
To ensure clarity and reproducibility of the intra-database protocol, we report the subject-disjoint train--test statistics used for FV-300, FV-USM, and MMCBNU in Table~\ref{tab:intra_split_stats}. The table reports the identity-level development/test partition, together with the number of train and test images used within the development identities.

For each dataset and statistical seed, identities are first sorted and partitioned at the \textbf{subject/identity level}. The first 80\% of identities are used for model development, and the remaining 20\% of identities are held out entirely for intra-database open-set testing. Thus, no subject or finger identity used for final open-set testing appears during model development.

Within the development identities, the dataset-provided train/test image folders are used for optimisation and validation/model selection, respectively. This image-level train/test split is not the source of open-set subject disjointness; subject disjointness is enforced by excluding the held-out 20\% identities from all model-development steps. During the final intra-database evaluation, embeddings are extracted only for the held-out identities and genuine/impostor scores are computed from within-identity and cross-identity image pairs among those unseen subjects.

For FV-300, the reported counts are approximate because 124 low-quality images were removed before partitioning. Despite these variations, the same subject-disjoint development/test splitting strategy is applied consistently across all intra-database experiments and all learned methods.





\begin{table}[ht!]
\centering
\scriptsize
\resizebox{0.5\textwidth}{!}{
\begin{tabular}{lcccc}
\hline
Dataset & Total IDs & Train / Val. IDs & Train / Val. imgs & Held-out IDs \\
\hline
FV-300 & 300 & 240 & 16,536 / 4,954 & 60 \\
MMCBNU & 600 & 480 & 2,400 / 2,400 & 120 \\
FV-USM & 492 & 393 & 2,358 / 2,358 & 99 \\
\hline
\end{tabular}}
\caption{Identity-level subject-disjoint train--test statistics for the intra-database open-set experiments. The first 80\% of sorted identities are used for model development, and the remaining identities are held out entirely for open-set testing. The development image counts report the dataset-provided train/test folders used for optimisation and validation/model selection within the development identities.}

\label{tab:intra_split_stats}
\end{table}

\section{t-SNE Visualisation and Embedding Analysis}
\label{sec:tsne_analysis}

To further analyse the structure of the learned embedding space, we visualise the features extracted by OpenVeinNet using t-SNE across five datasets. For each dataset, we present three complementary views: (i) best 50 classes, (ii) worst 50 classes, and (iii) all classes. The best and worst classes are selected based on class-wise verification performance, allowing us to inspect embedding separability under favourable and challenging conditions.

\begin{figure*}[htbp!]
    \centering
    \begin{subfigure}[b]{0.3\textwidth}
        \centering
        \includegraphics[width=\textwidth]{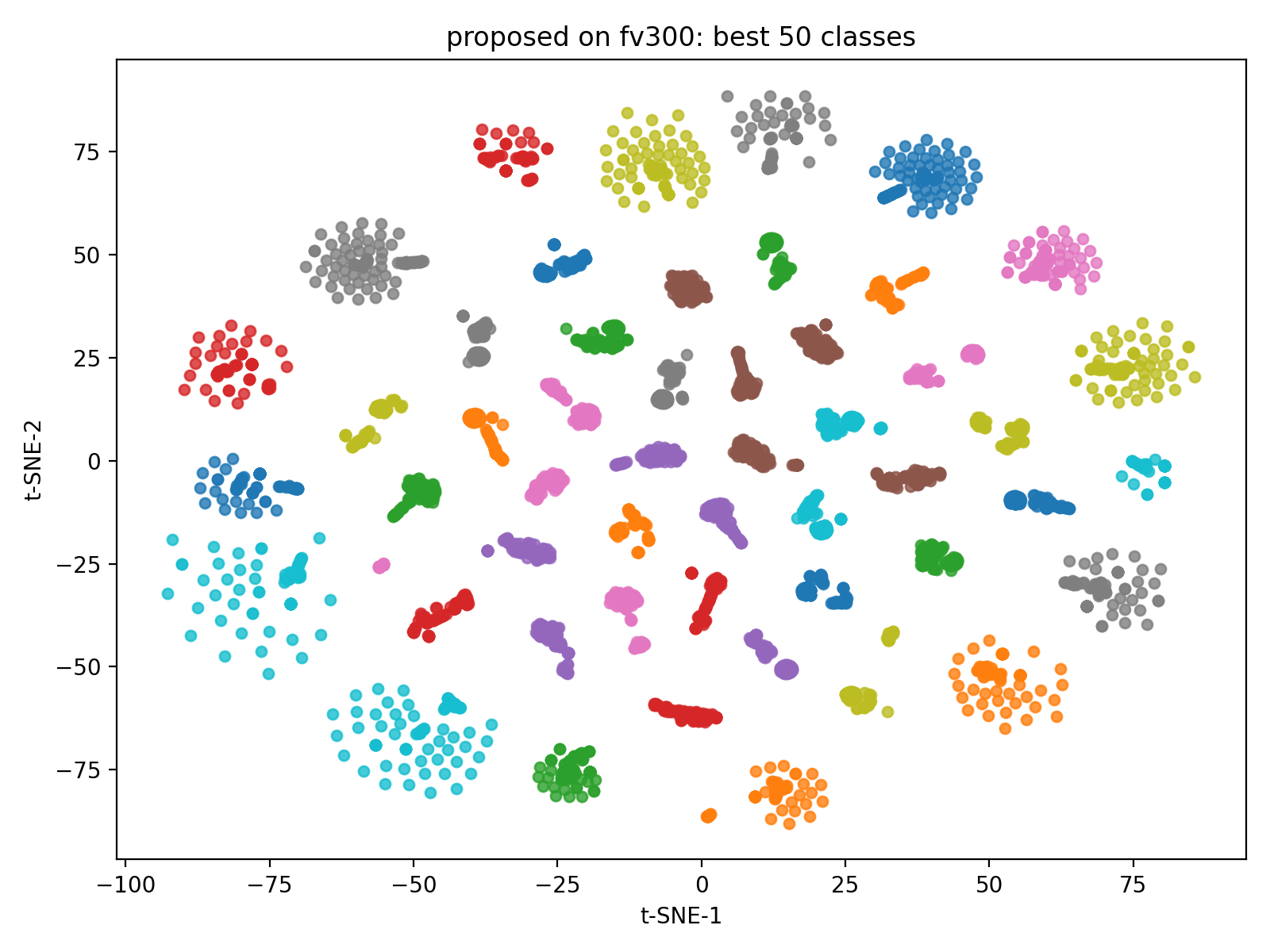}
        \caption{Best 50 classes}
        \label{fig:fv300_best}
    \end{subfigure}
    \hfill 
    \begin{subfigure}[b]{0.3\textwidth}
        \centering
        \includegraphics[width=\textwidth]{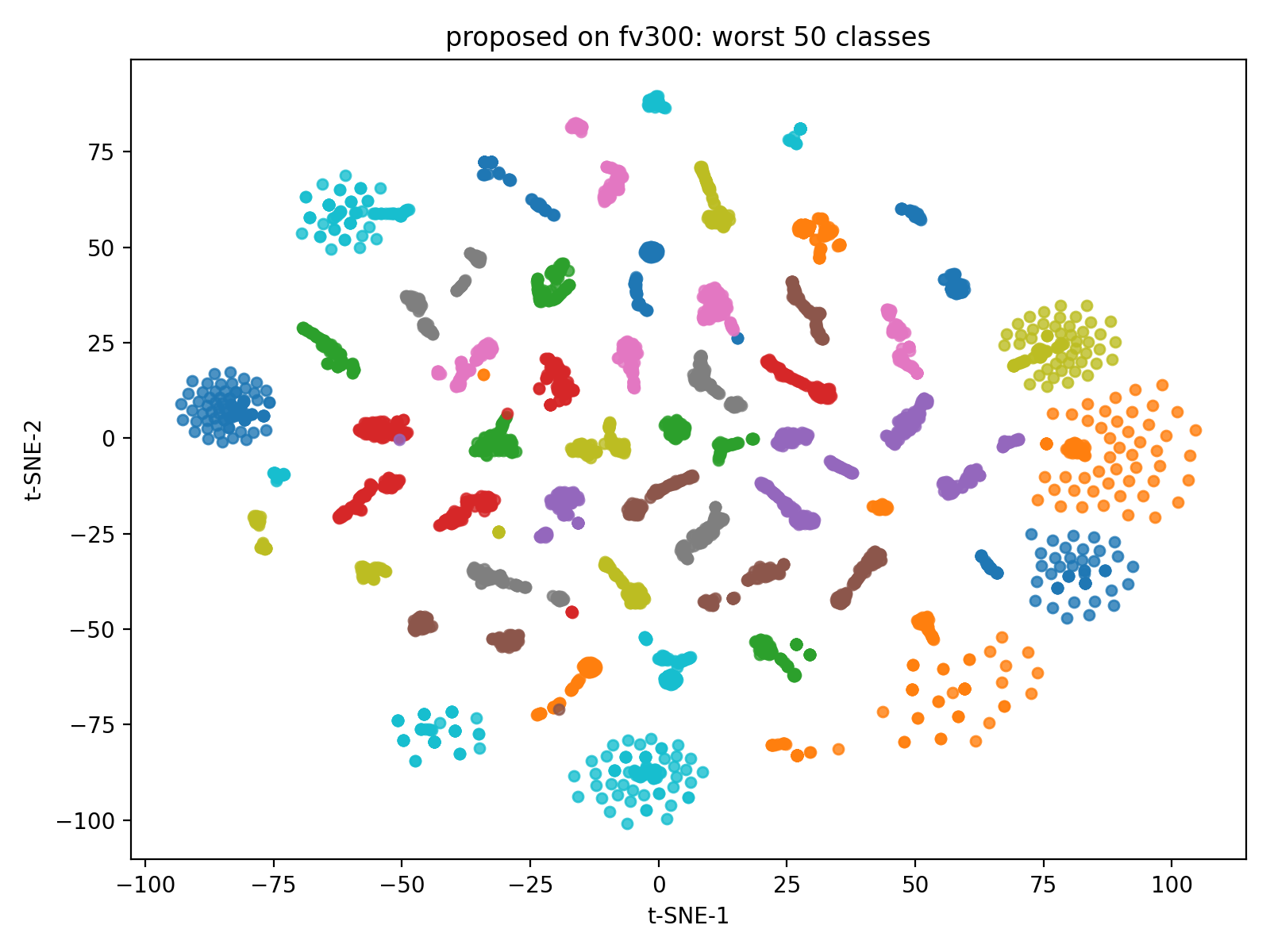}
        \caption{Worst 50 classes}
        \label{fig:fv300_worst}
    \end{subfigure}
    \hfill
    \begin{subfigure}[b]{0.3\textwidth}
        \centering
        \includegraphics[width=\textwidth]{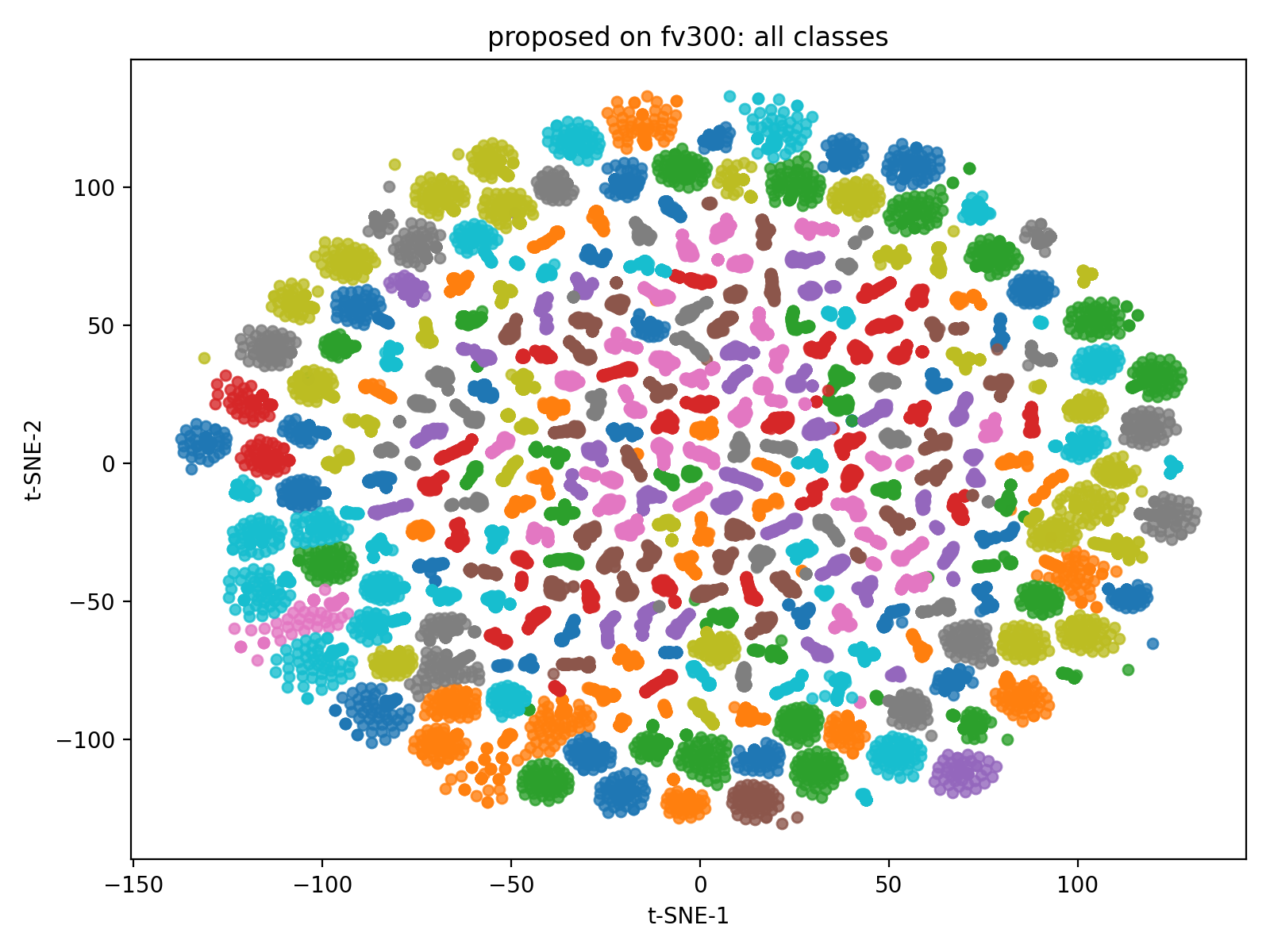}
        \caption{All Classes}
        \label{fig:fv300_all}
    \end{subfigure}
    \caption{FV-300 Database t-SNE visualisation of OpenVeinNet embeddings: (a) best 50 classes, (b) worst 50 classes, and (c) all classes.}
    \label{fig:fv300}
\end{figure*}

\textbf{FV-300 (see Fig.~\ref{fig:fv300}).}
The t-SNE visualisation for FV-300 shows well-defined and compact clusters for the best 50 classes, indicating strong intra-class compactness and clear inter-class separation. Even in the worst 50 classes, most identities remain reasonably separable, although slight overlaps appear in boundary regions. The all-class distribution exhibits a structured global arrangement, suggesting that the learned embeddings preserve identity-level discriminability across the dataset. This behaviour is consistent with the strong performance observed in the $BCDE \rightarrow A$ protocol, where FV-300 provides high-quality images and a large number of samples per identity.

\begin{figure*}[t] 
    \centering
    \begin{minipage}[b]{0.31\textwidth}
        \centering
        \includegraphics[width=\textwidth]{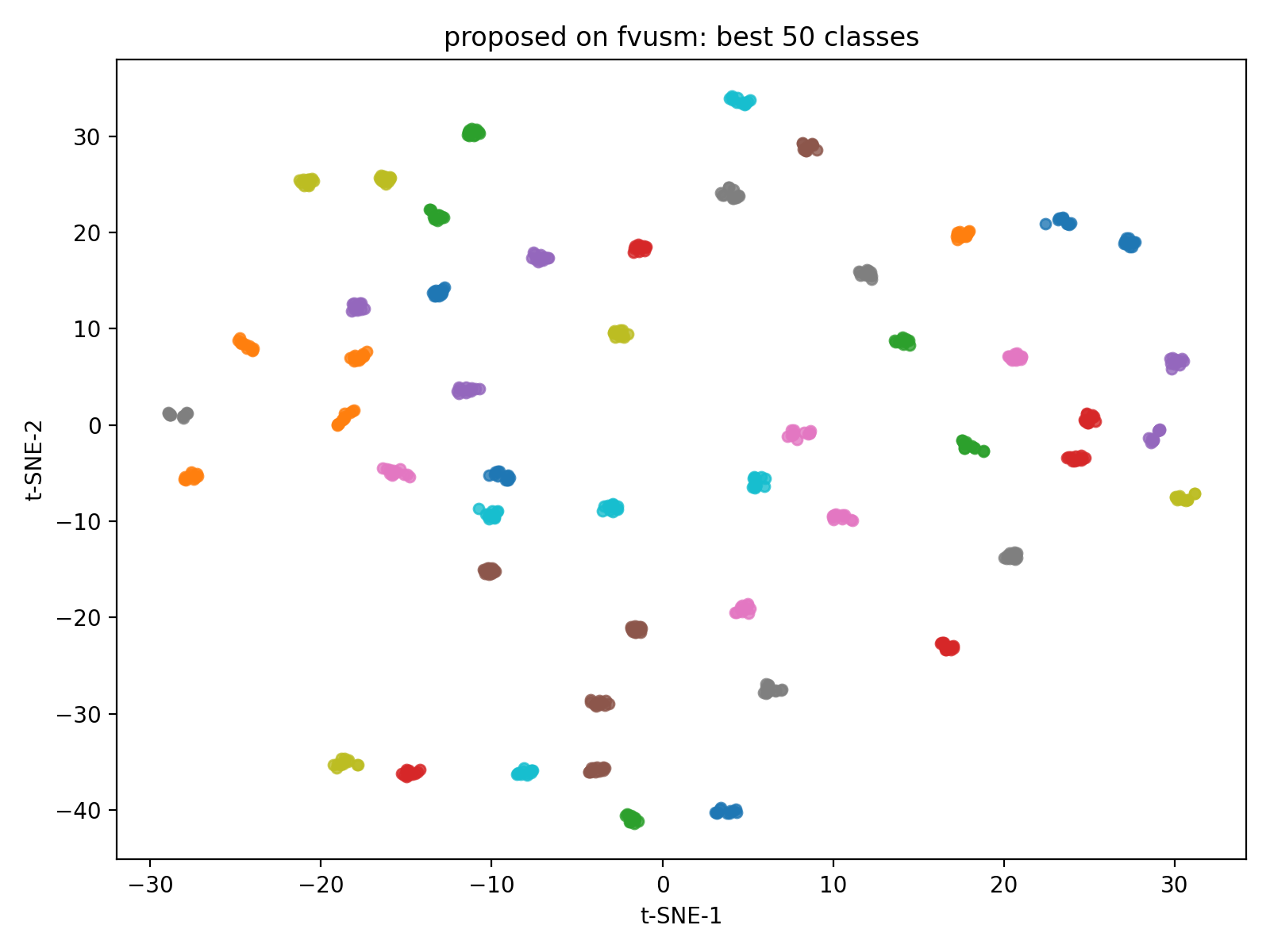}
        \caption{Best 50 classes}
        \label{fig:fvusm_best}
    \end{minipage}
    \hfill 
    \begin{minipage}[b]{0.31\textwidth}
        \centering
        \includegraphics[width=\textwidth]{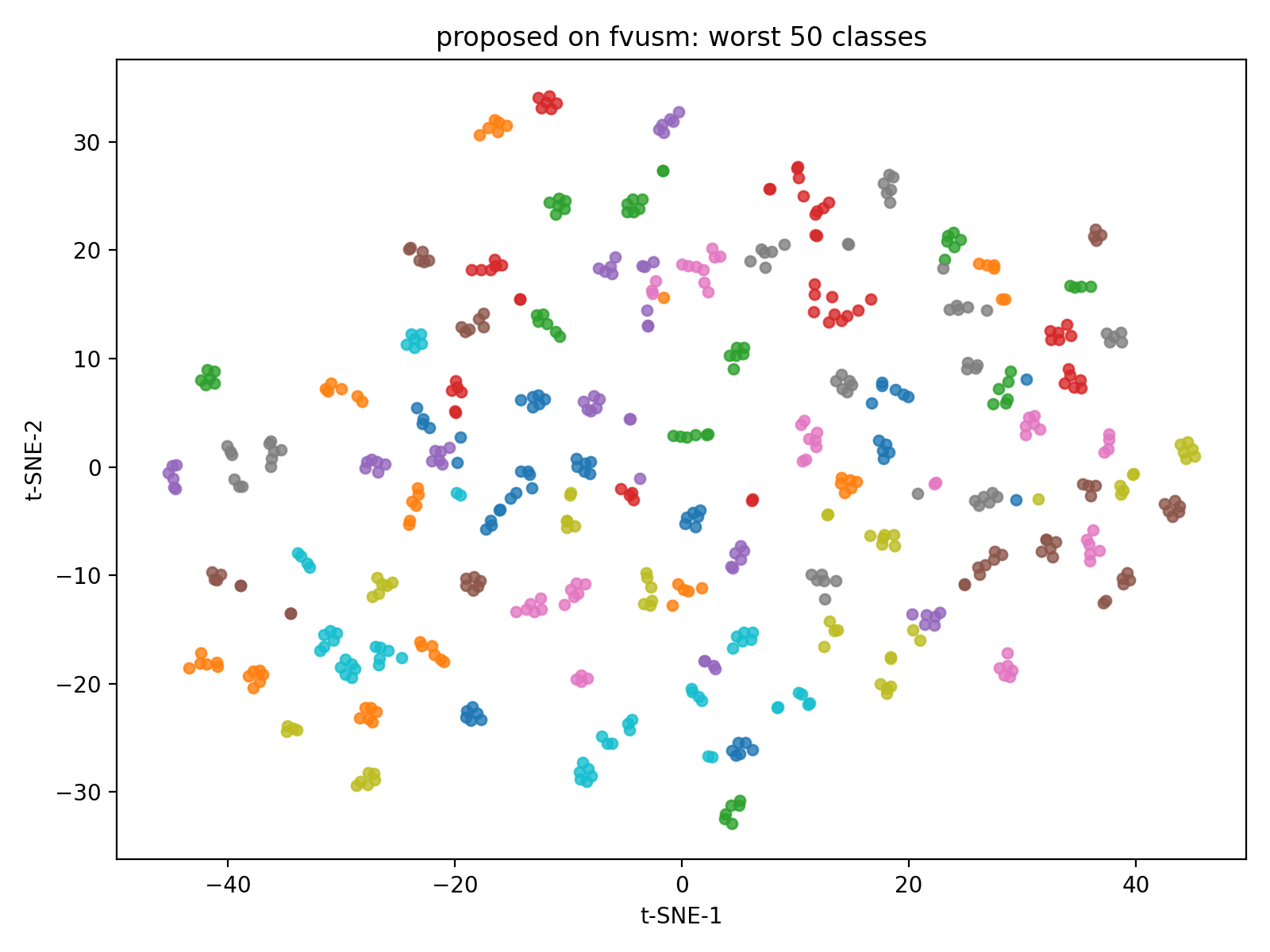}
        \caption{Worst 50 classes}
        \label{fig:fvusm_worst}
    \end{minipage}
    \hfill
    \begin{minipage}[b]{0.31\textwidth}
        \centering
        \includegraphics[width=\textwidth]{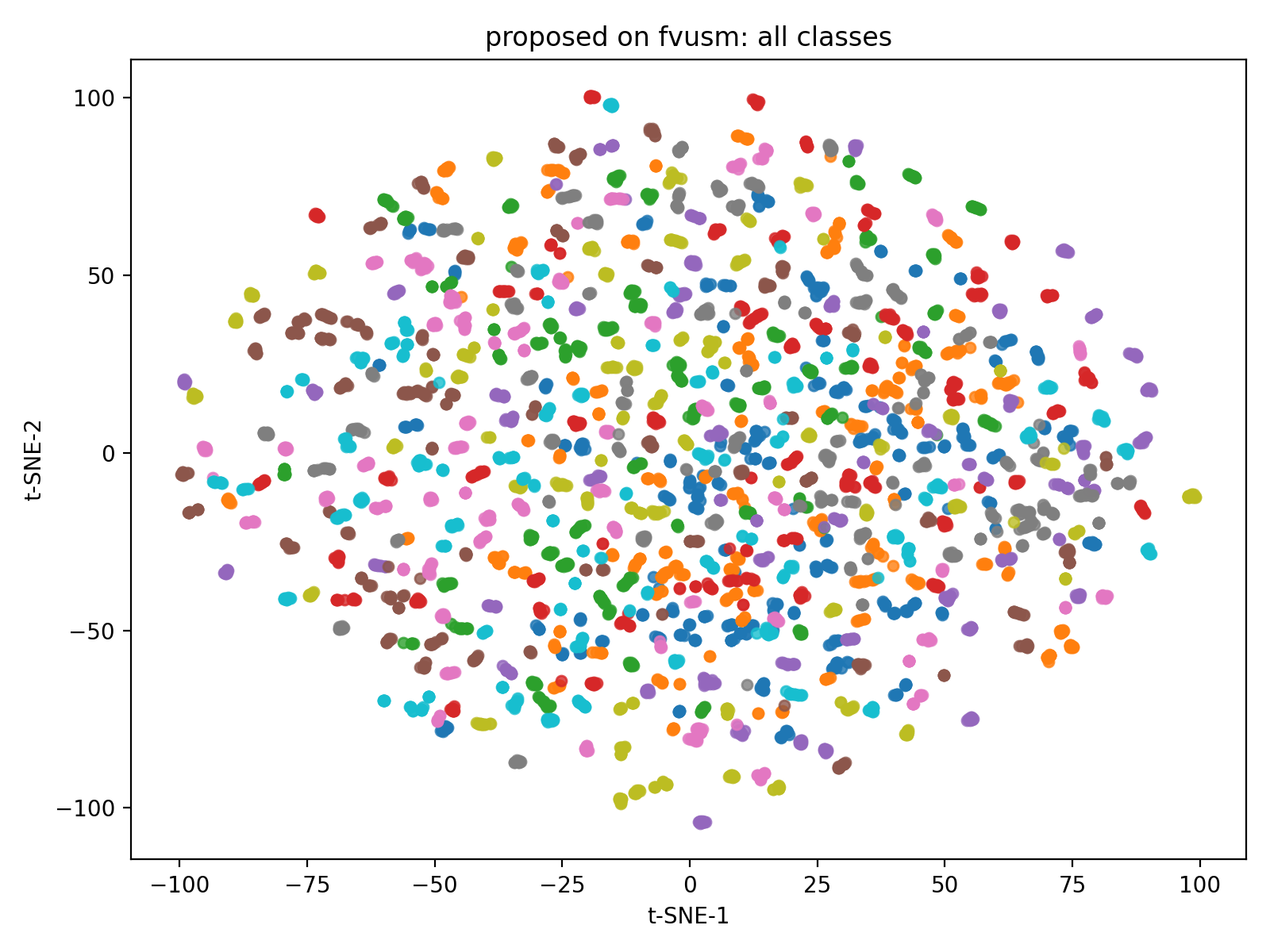}
        \caption{All Classes}
        \label{fig:fvusm_alll}
    \end{minipage}
    
    \vspace{2mm}
    \caption{FV-USM database t-SNE visualisation of OpenVeinNet embeddings: (a) best 50 classes, (b) worst 50 classes, and (c) all classes.}
    \label{fig:fvusm}
\end{figure*}

\textbf{FV-USM (see Fig.~\ref{fig:fvusm}).}
For FV-USM, the best-class embeddings remain relatively compact but are more scattered compared to FV-300. The worst 50 classes show noticeable overlap and reduced cluster density, indicating increased intra-class variation and inter-class ambiguity. The all-class visualisation reveals a less structured global distribution, reflecting the moderate image quality and cross-session variability present in this dataset. Despite this, the presence of local grouping structures suggests that OpenVeinNet retains meaningful identity information under domain shift.

\begin{figure*}[t]
    \centering
    \begin{minipage}[b]{0.31\textwidth}
        \centering
        \includegraphics[width=\textwidth]{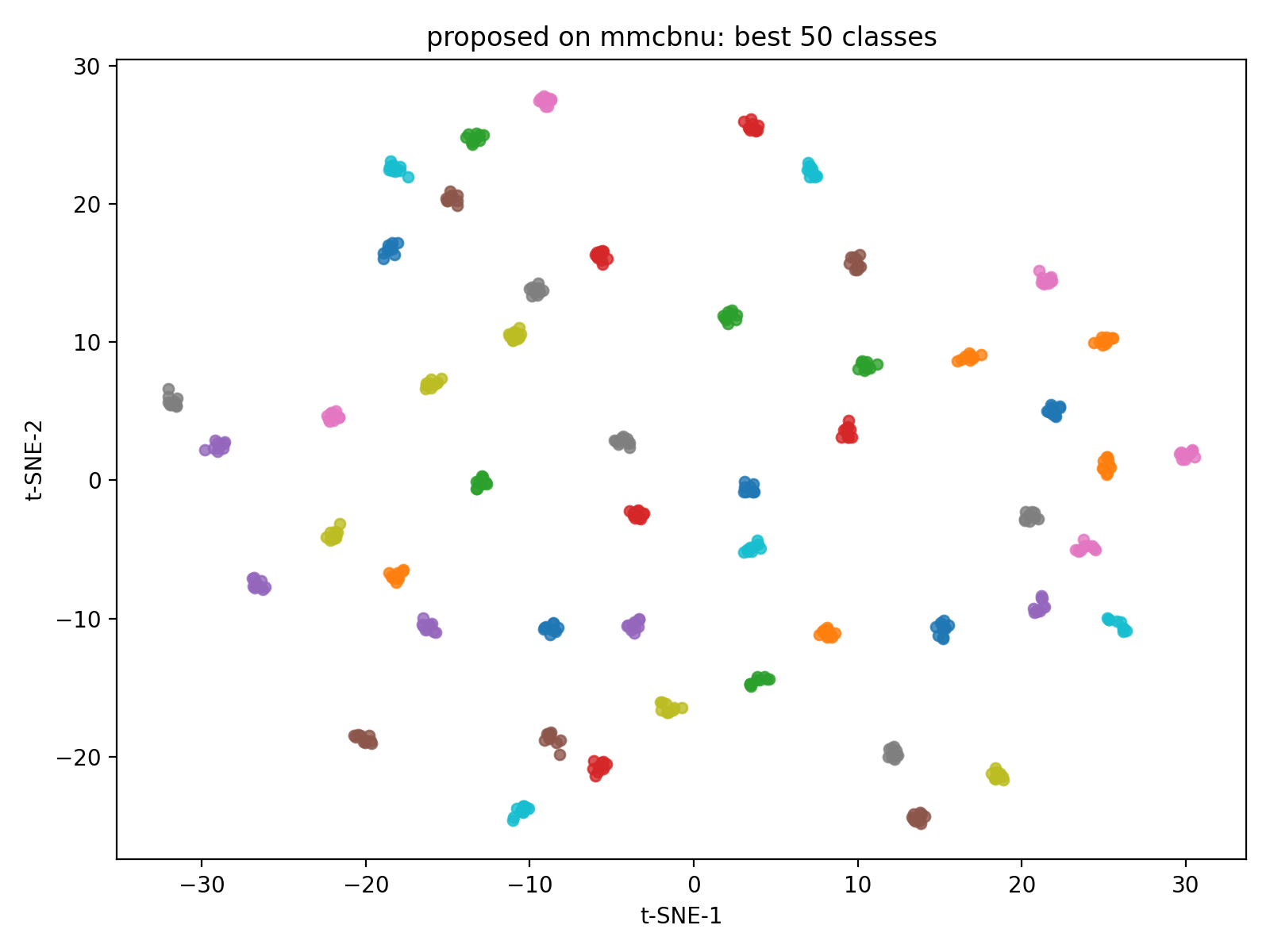}
        \caption{Best 50 classes}
        \label{fig:mmcbnu_best}
    \end{minipage}
    \hfill 
    \begin{minipage}[b]{0.31\textwidth}
        \centering
        \includegraphics[width=\textwidth]{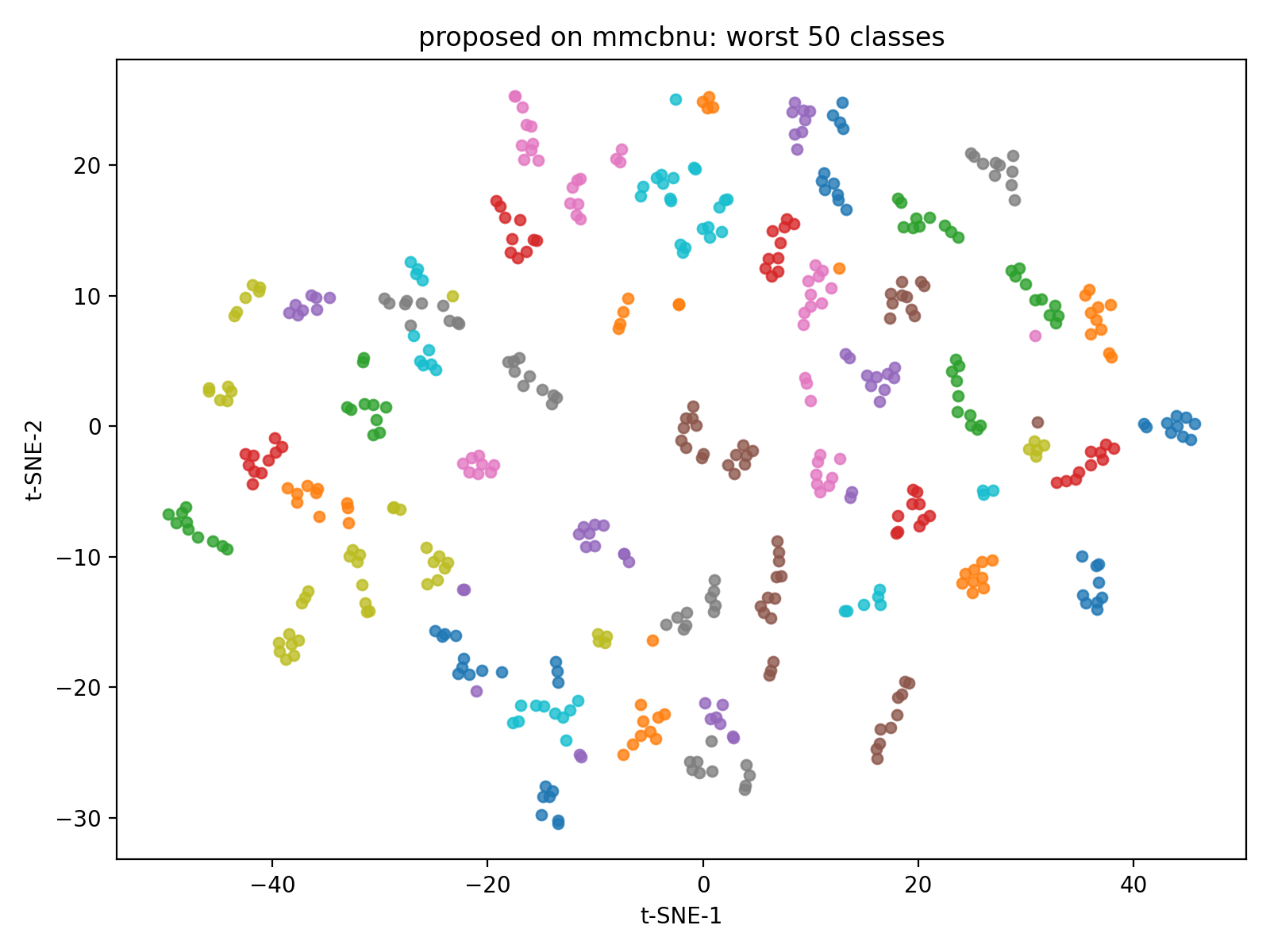}
        \caption{Worst 50 classes}
        \label{fig:mmcbnu_worst}
    \end{minipage}
    \hfill
    \begin{minipage}[b]{0.31\textwidth}
        \centering
        \includegraphics[width=\textwidth]{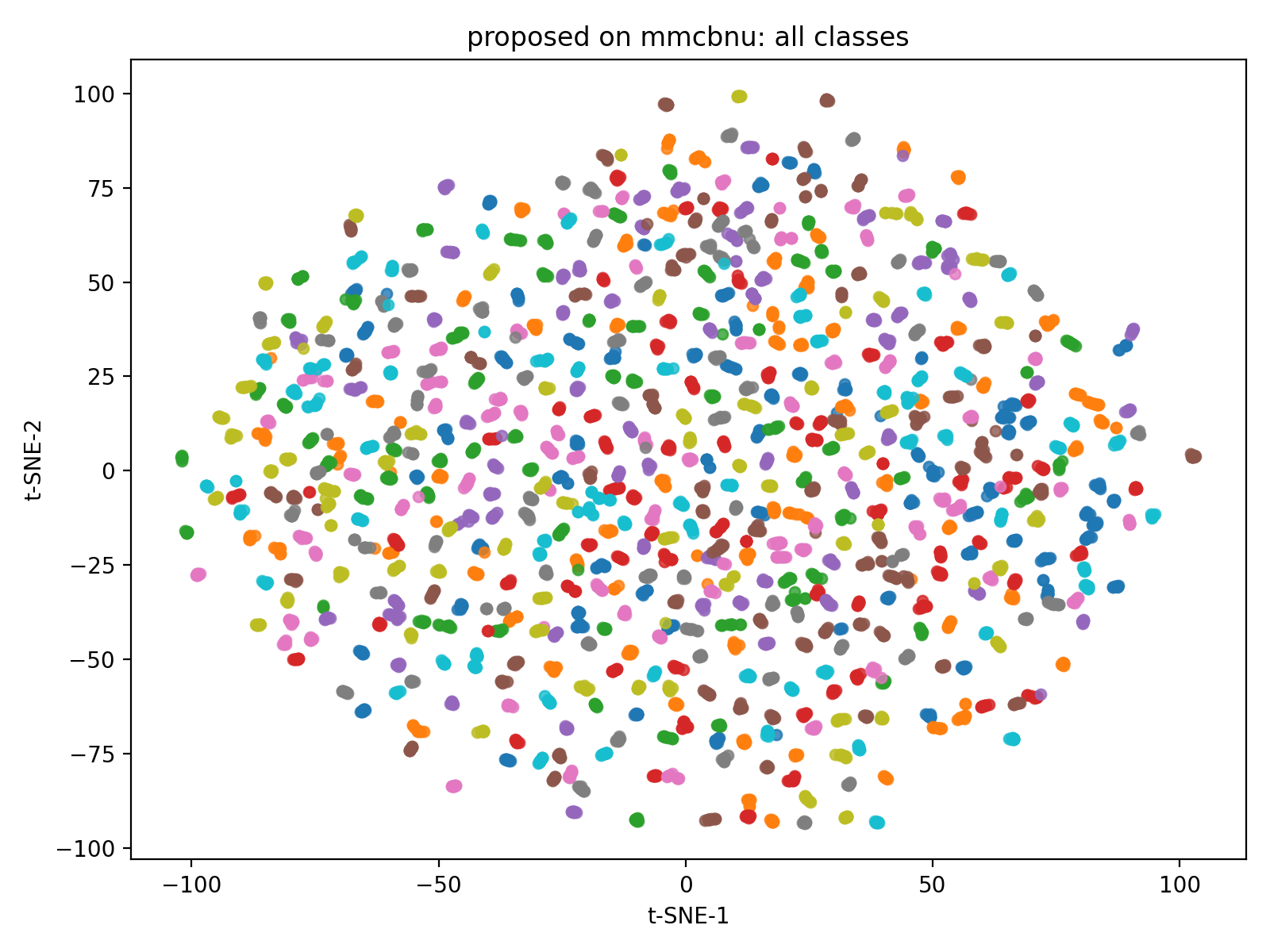}
        \caption{All Classes}
        \label{fig:mmcbnu_alll}
    \end{minipage}
    
    \vspace{2mm}
    \caption{MMCBNU database t-SNE visualisation of OpenVeinNet embeddings: (a) best 50 classes, (b) worst 50 classes, and (c) all classes.}
    \label{fig:mmcbnu}
\end{figure*}

\textbf{MMCBNU (see Fig.~\ref{fig:mmcbnu}).}
The MMCBNU visualisation shows moderate cluster separability in the best 50 classes, while the worst classes exhibit increased dispersion and partial mixing. The all-class view reveals a relatively uniform distribution of embeddings, indicating that class boundaries are less sharply defined compared to FV-300. This behaviour can be attributed to variations in acquisition conditions and subject diversity. Nevertheless, identifiable cluster structures are still present, supporting the model’s ability to generalise across heterogeneous data.

\begin{figure*}[t]
    \centering
    \begin{minipage}[b]{0.31\textwidth}
        \centering
        \includegraphics[width=\textwidth]{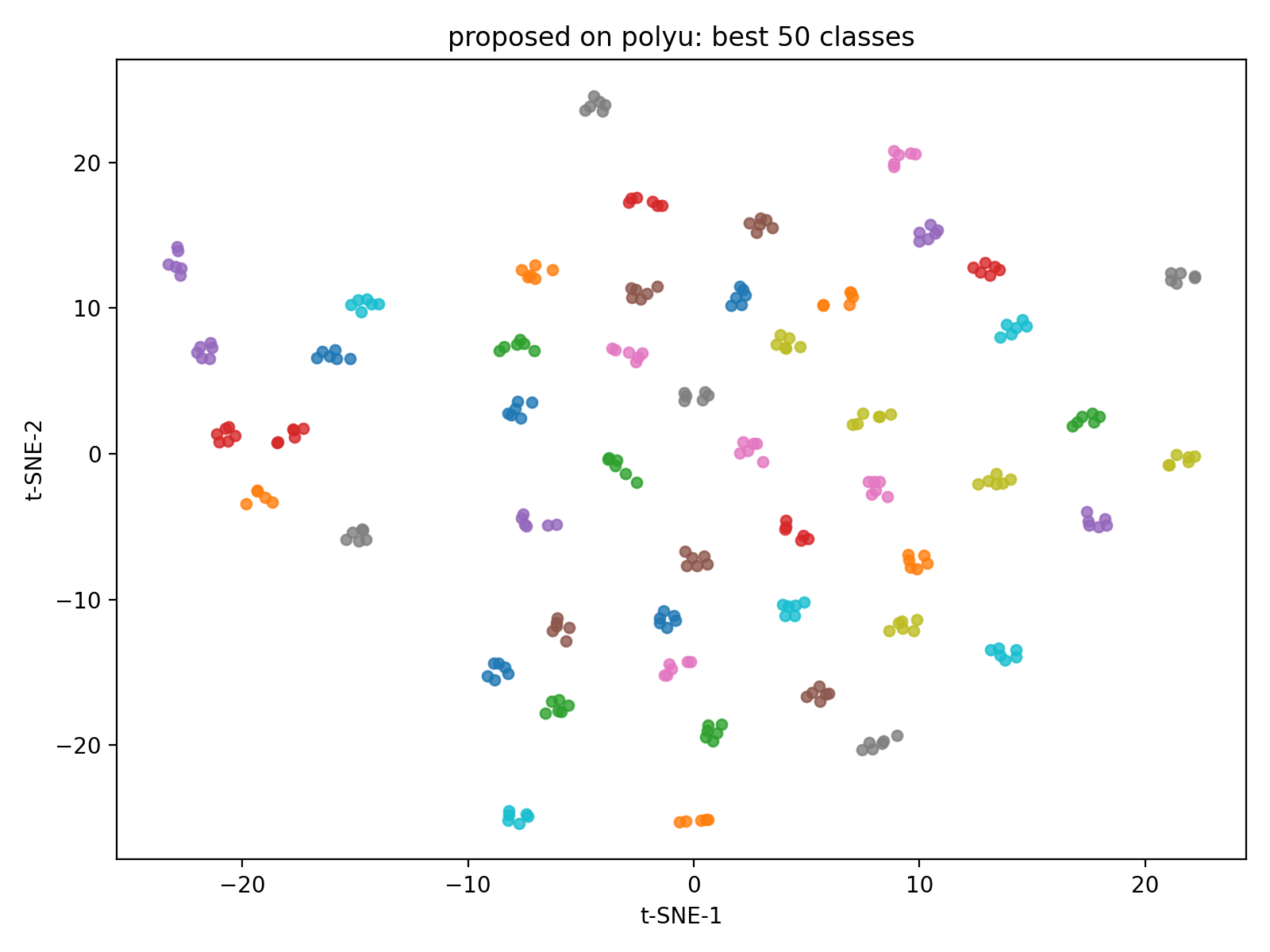}
        \caption{Best 50 classes}
        \label{fig:polyu_best}
    \end{minipage}
    \hfill 
    \begin{minipage}[b]{0.31\textwidth}
        \centering
        \includegraphics[width=\textwidth]{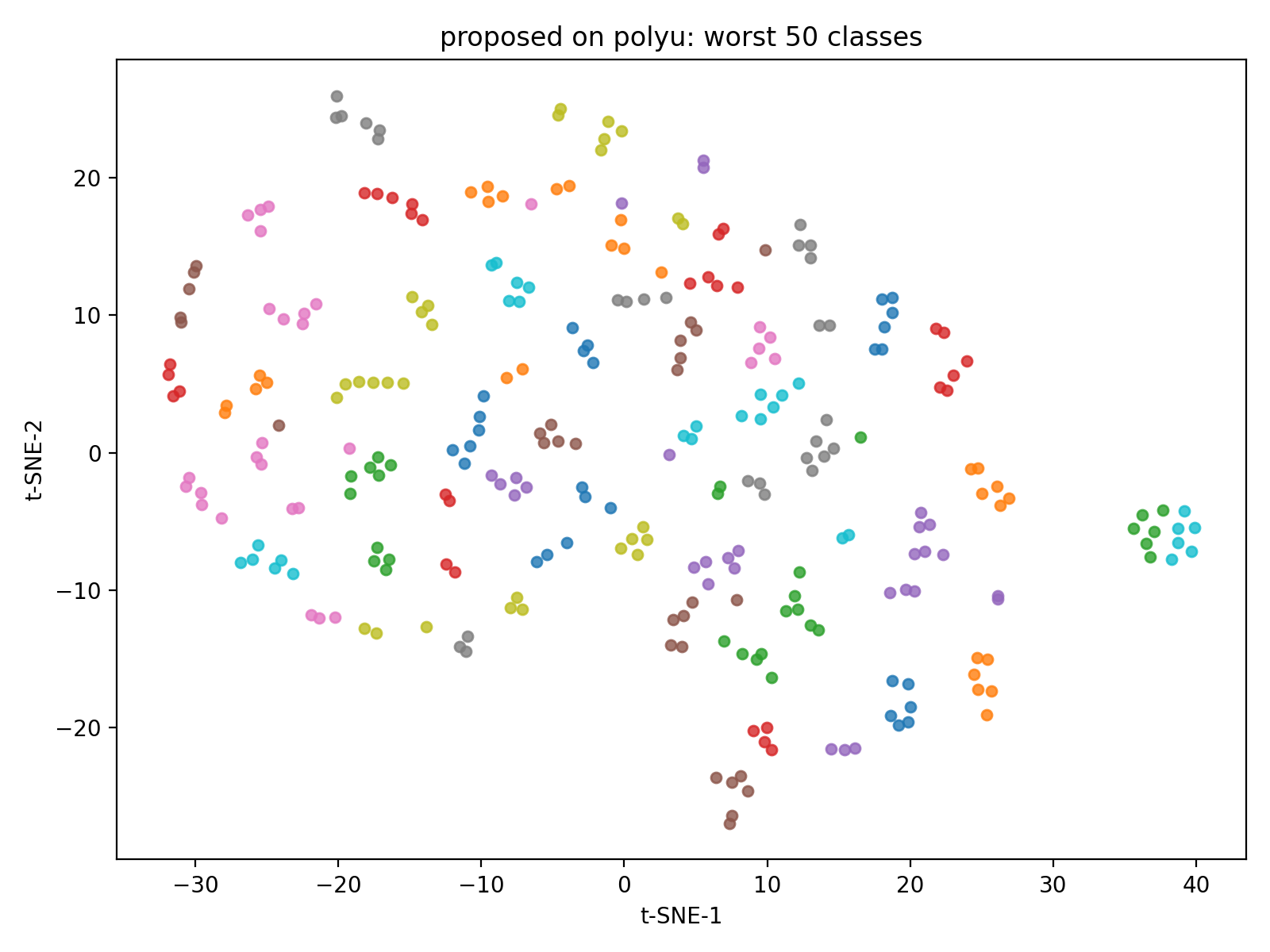}
        \caption{Worst 50 classes}
        \label{fig:polyu_worst}
    \end{minipage}
    \hfill
    \begin{minipage}[b]{0.31\textwidth}
        \centering
        \includegraphics[width=\textwidth]{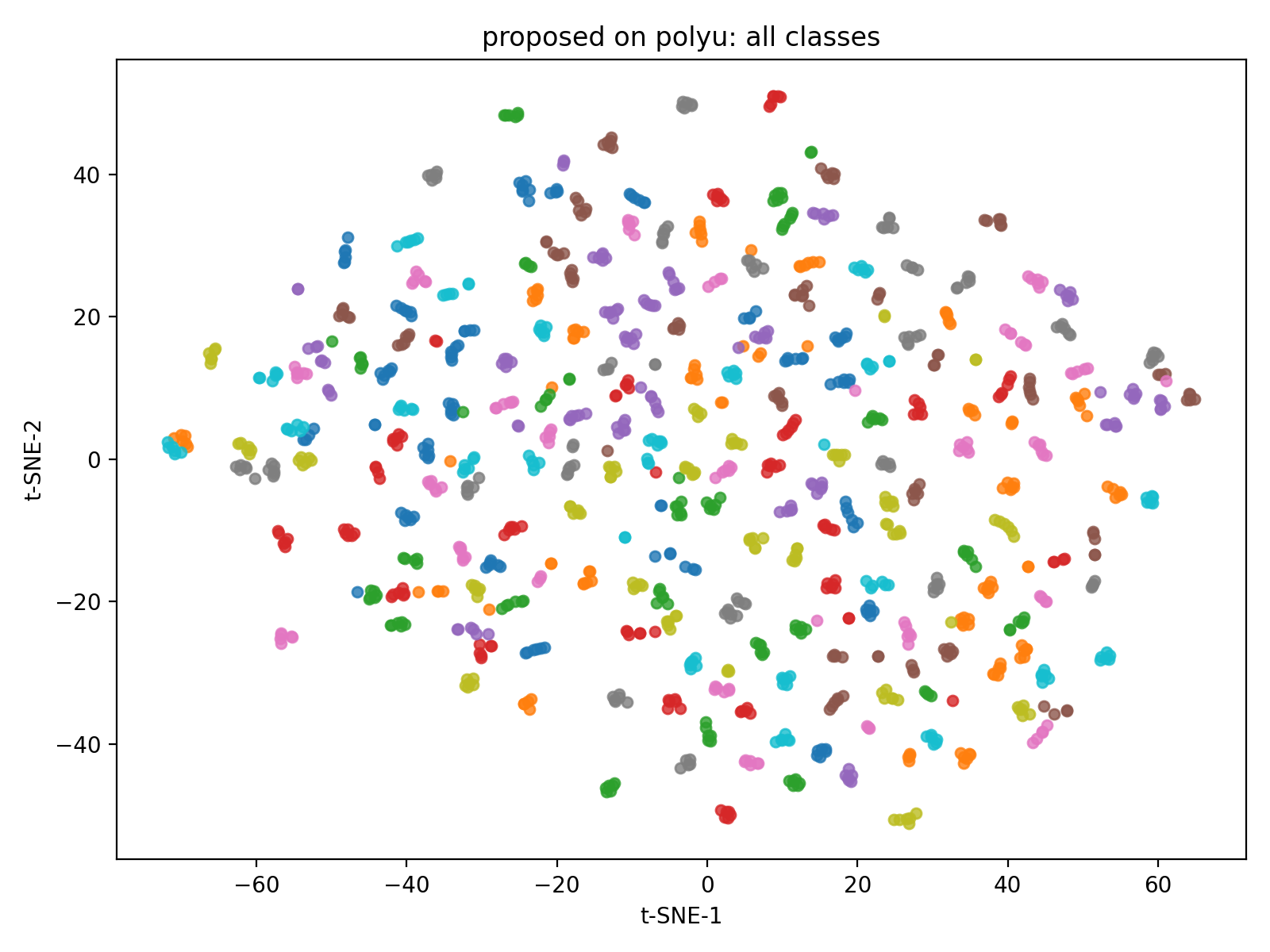}
        \caption{All Classes}
        \label{fig:polyu_alll}
    \end{minipage}
    
    \vspace{2mm}
    \caption{PolyU database t-SNE visualisation of OpenVeinNet embeddings: (a) best 50 classes, (b) worst 50 classes, and (c) all classes.}
    \label{fig:polyu}
\end{figure*}

\textbf{PolyU (see Fig.~\ref{fig:polyu}).}
In the PolyU dataset, the best 50 classes show distinguishable clusters, though with reduced compactness compared to FV-300. The worst classes display significant overlap and irregular cluster shapes, indicating difficulty in separating certain identities. The all-class distribution appears more diffused, reflecting the unconstrained acquisition setup and long-term variability in the dataset. These observations are consistent with the moderate performance trends reported in the main experiments.

\begin{figure*}[t]
    \centering
    \begin{minipage}[b]{0.31\textwidth}
        \centering
        \includegraphics[width=\textwidth]{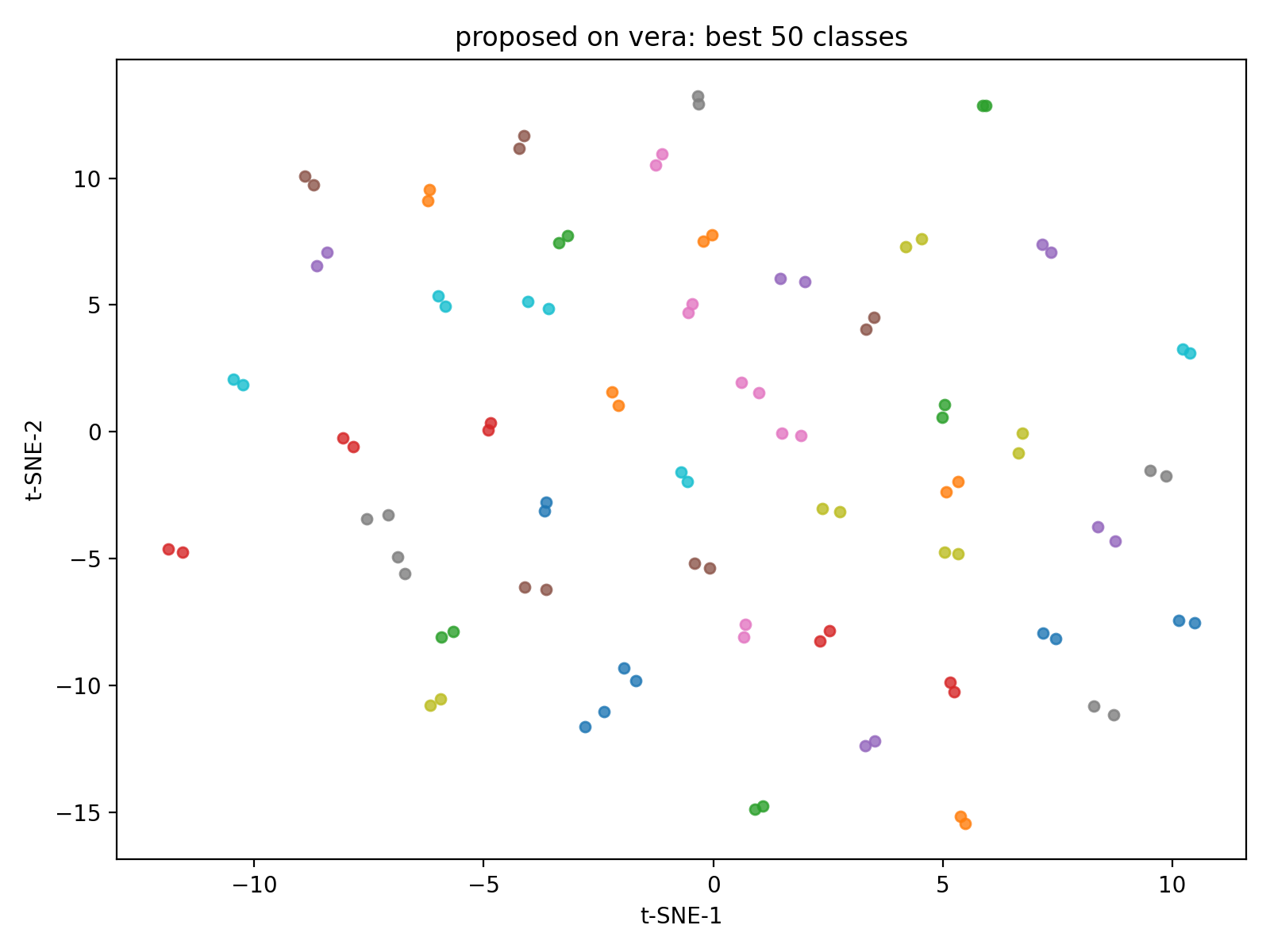}
        \caption{Best 50 classes}
        \label{fig:vera_best}
    \end{minipage}
    \hfill 
    \begin{minipage}[b]{0.31\textwidth}
        \centering
        \includegraphics[width=\textwidth]{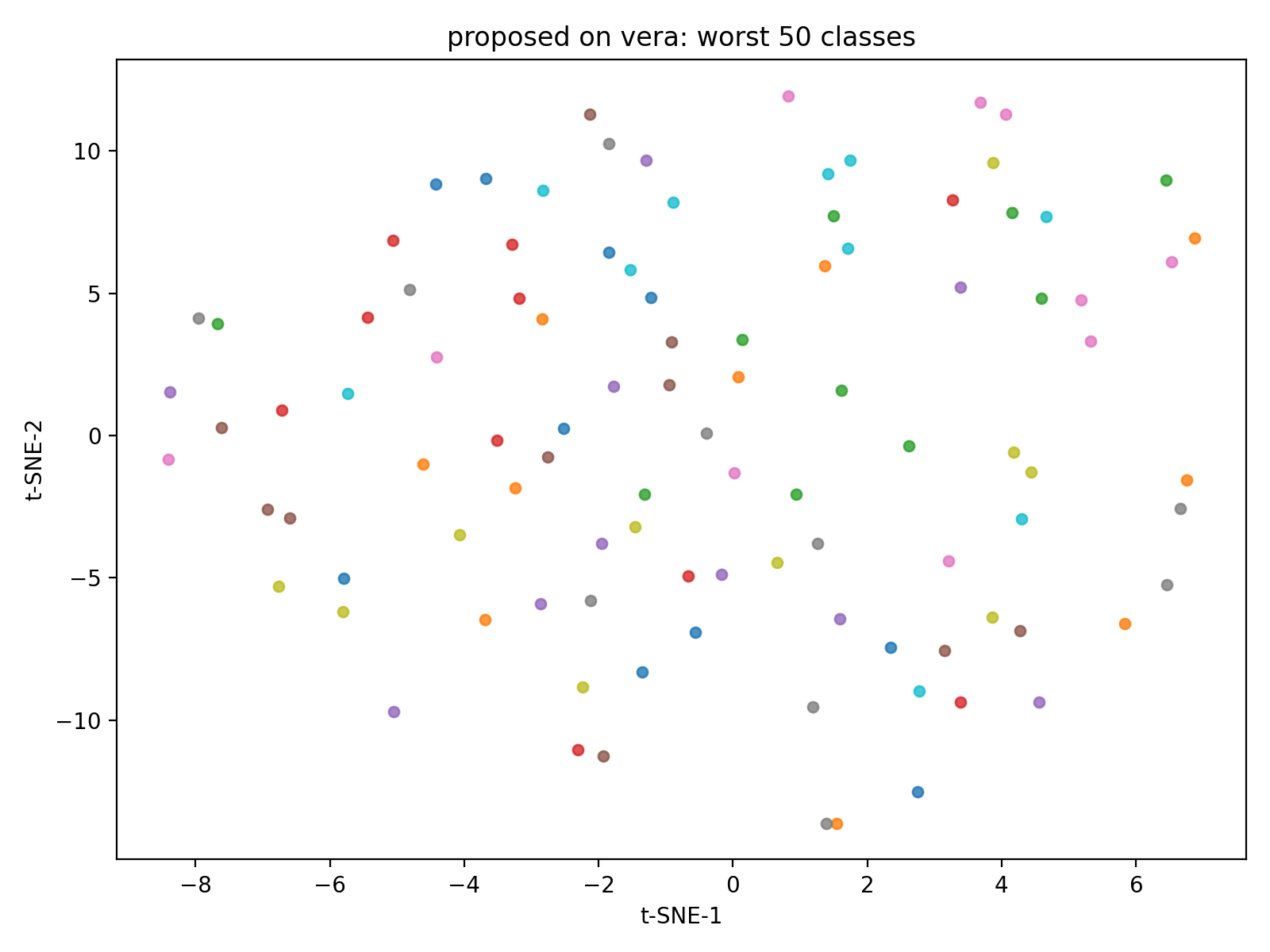}
        \caption{Worst 50 classes}
        \label{fig:vera_worst}
    \end{minipage}
    \hfill
    \begin{minipage}[b]{0.31\textwidth}
        \centering
        \includegraphics[width=\textwidth]{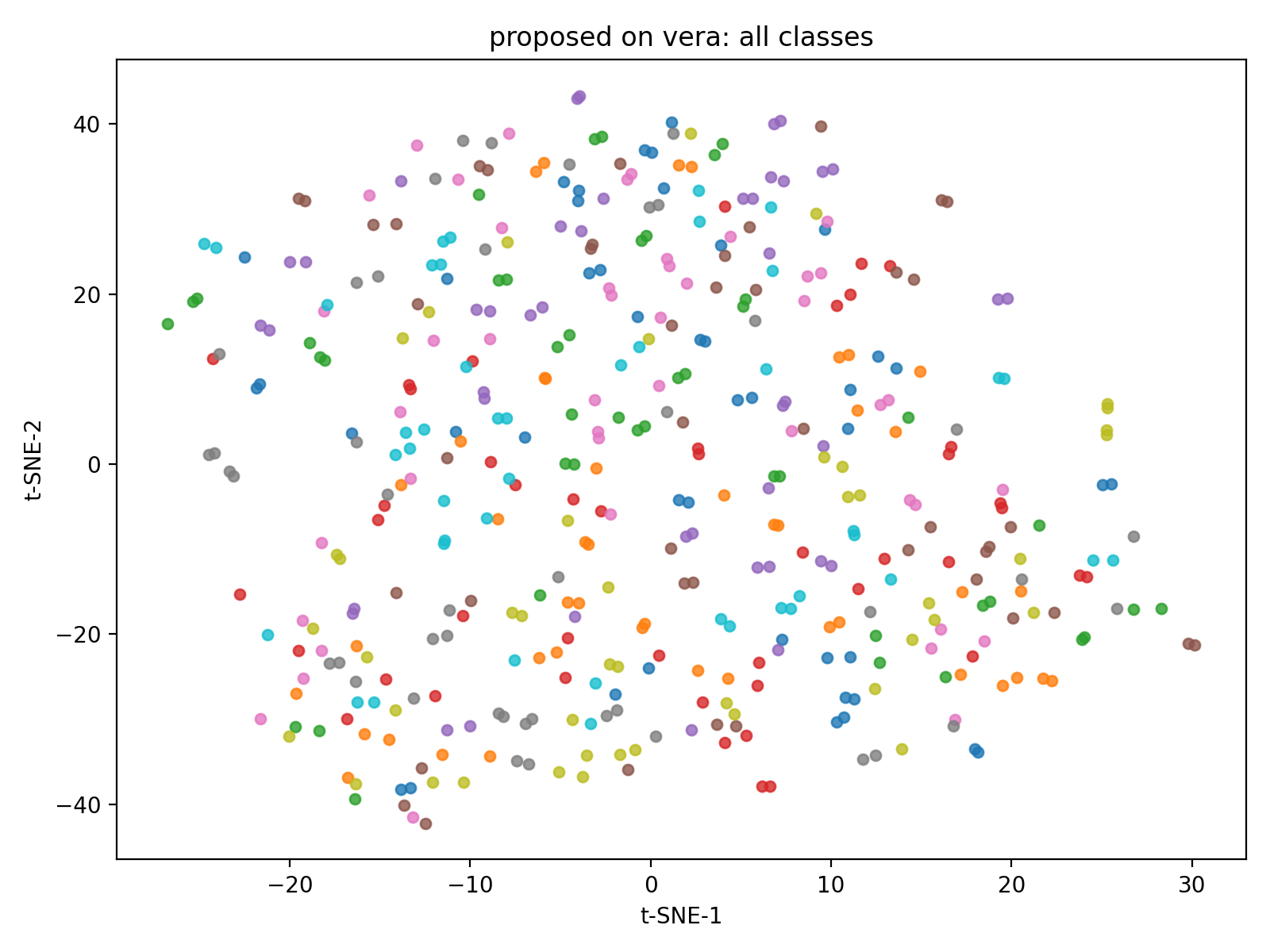}
        \caption{All Classes}
        \label{fig:vera_alll}
    \end{minipage}
    
    \vspace{2mm}
    \caption{VERA database t-SNE visualisation of OpenVeinNet embeddings: (a) best 50 classes, (b) worst 50 classes, and (c) all classes.}
    \label{fig:vera}
\end{figure*}

\textbf{VERA (see Fig.~\ref{fig:vera}).}
The VERA dataset presents the most challenging embedding structure. Even in the best 50 classes, clusters are sparse and less compact, while the worst classes show substantial overlap and poor separability. The all-class visualisation exhibits a highly scattered distribution with minimal cluster structure. This is expected due to the extremely limited number of samples per identity (only two images) and higher acquisition variability. The t-SNE results therefore provide strong qualitative support for the performance degradation observed in the $ABCD \rightarrow E$ protocol.

\textbf{Overall analysis:}
Across all datasets, the t-SNE visualisations reveal a consistent pattern: datasets with higher image quality and more samples per identity (e.g., FV-300) produce well-separated and compact clusters, while low-sample and high-variability datasets (e.g., VERA) result in dispersed and overlapping embeddings. Importantly, even in challenging scenarios, the proposed OpenVeinNet maintains partial cluster structures, indicating that the learned representation remains discriminative.

These findings complement the quantitative results and dataset-quality analysis presented in the main paper. Together, they confirm that the performance variations across datasets are primarily driven by data characteristics rather than instability in the proposed model.

\section{DET Curve Analysis}
We further analyse the verification behaviour using DET curves under both half and full evaluation protocols (Fig.~18 and Fig.~19). Across all datasets, the proposed method consistently achieves curves closer to the origin, indicating a favourable trade-off between false match rate (FMR) and false non-match rate (FNMR).

On FV-300, all methods perform strongly due to high data quality; however, the proposed method maintains lower FNMR, particularly in the low-FMR regime, which is critical for secure applications. On FV-USM and MMCBNU, the performance gap becomes more pronounced, with OpenVeinNet showing consistently lower FNMR across most operating regions, reflecting improved robustness to intra-class variation and acquisition differences. 

For PolyU, the DET curves exhibit greater dispersion across methods, yet the proposed approach retains a clear advantage, especially at moderate FMR values. VERA remains the most challenging dataset, where all methods degrade; nevertheless, OpenVeinNet maintains comparatively better performance and exhibits a more gradual degradation, indicating improved generalisation under low-sample and high-variability conditions.

The trends are consistent between half and full protocols, with the full protocol showing slight improvements due to increased enrolment data. Overall, the DET analysis confirms that OpenVeinNet provides stable and favourable operating characteristics across datasets, particularly in low-FMR regimes relevant to security-critical deployments.

\begin{figure*}[t]
    \centering
    \begin{subfigure}[b]{0.31\textwidth}
        \centering
        \includegraphics[width=\textwidth]{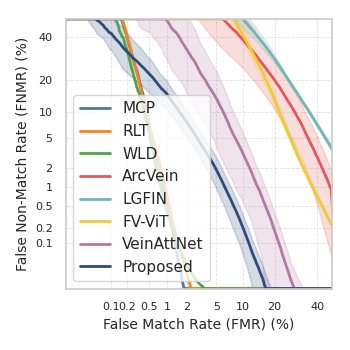}
        \caption{FV-300}
        \label{fig:det_half_fv300}
    \end{subfigure}
    \hfill
    \begin{subfigure}[b]{0.31\textwidth}
        \centering
        \includegraphics[width=\textwidth]{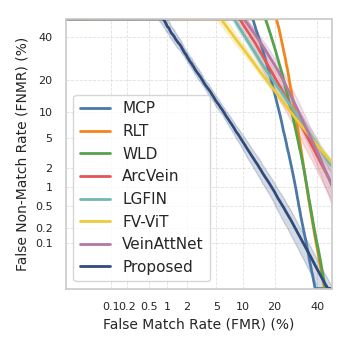}
        \caption{FV-USM}
        \label{fig:det_half_fvusm}
    \end{subfigure}
    \hfill
    \begin{subfigure}[b]{0.31\textwidth}
        \centering
        \includegraphics[width=\textwidth]{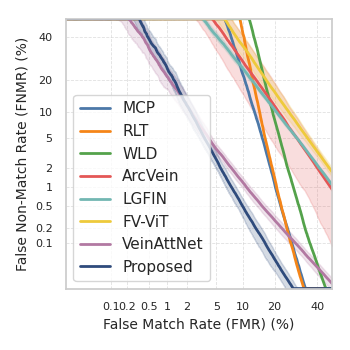}
        \caption{MMCBNU}
        \label{fig:det_half_mmcbnu}
    \end{subfigure}

    \vspace{2mm}

    \hfill
    \begin{subfigure}[b]{0.31\textwidth}
        \centering
        \includegraphics[width=\textwidth]{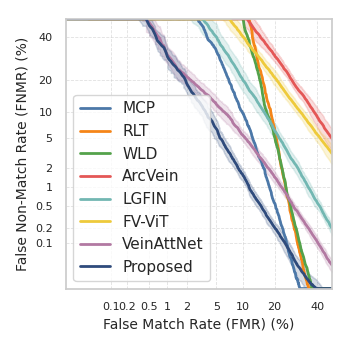}
        \caption{PolyU}
        \label{fig:det_half_polyu}
    \end{subfigure}
    \hfill
    \begin{subfigure}[b]{0.31\textwidth}
        \centering
        \includegraphics[width=\textwidth]{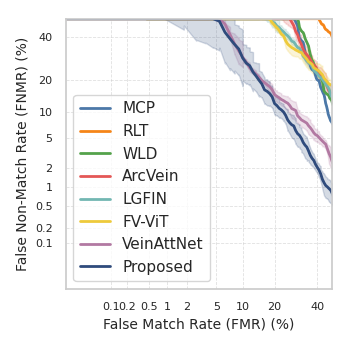}
        \caption{VERA}
        \label{fig:det_half_vera}
    \end{subfigure}
    \hfill

    \caption{DET curves for the half evaluation protocol across the five datasets: (a) FV-300, (b) FV-USM, (c) MMCBNU, (d) PolyU, and (e) VERA.}
    \label{fig:det_half_group}
\end{figure*}

\begin{figure*}[t]
    \centering
    \begin{subfigure}[b]{0.31\textwidth}
        \centering
        \includegraphics[width=\textwidth]{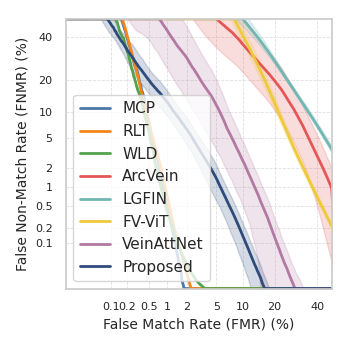}
        \caption{FV-300}
        \label{fig:det_full_fv300}
    \end{subfigure}
    \hfill
    \begin{subfigure}[b]{0.31\textwidth}
        \centering
        \includegraphics[width=\textwidth]{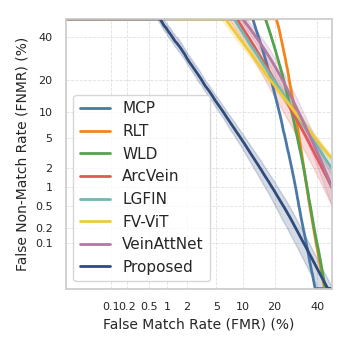}
        \caption{FV-USM}
        \label{fig:det_full_fvusm}
    \end{subfigure}
    \hfill
    \begin{subfigure}[b]{0.31\textwidth}
        \centering
        \includegraphics[width=\textwidth]{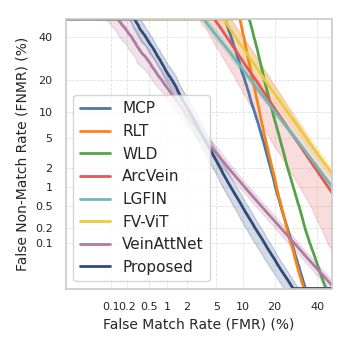}
        \caption{MMCBNU}
        \label{fig:det_full_mmcbnu}
    \end{subfigure}

    \vspace{2mm}

    \hfill
    \begin{subfigure}[b]{0.31\textwidth}
        \centering
        \includegraphics[width=\textwidth]{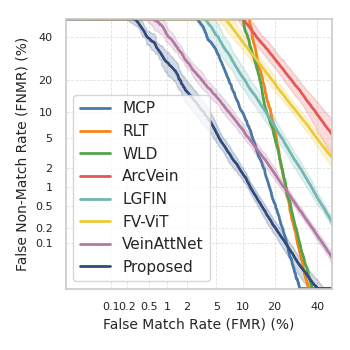}
        \caption{PolyU}
        \label{fig:det_full_polyu}
    \end{subfigure}
    \hfill
    \begin{subfigure}[b]{0.31\textwidth}
        \centering
        \includegraphics[width=\textwidth]{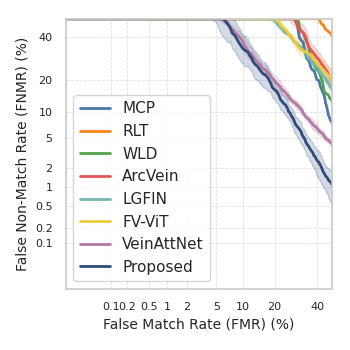}
        \caption{VERA}
        \label{fig:det_full_vera}
    \end{subfigure}
    \hfill

    \caption{DET curves for the full evaluation protocol across the five datasets: (a) FV-300, (b) FV-USM, (c) MMCBNU, (d) PolyU, and (e) VERA.}
    \label{fig:det_full_group}
\end{figure*}




\end{document}